\pdfoutput=1

\documentclass[11pt]{article}

\usepackage{ACL2023}

\usepackage{times}
\usepackage{latexsym}
\usepackage{hyperref}

\usepackage[T1]{fontenc}

\usepackage[utf8]{inputenc}

\usepackage{microtype}

\usepackage{inconsolata}

\usepackage{longtable}

\usepackage{xcolor}
\usepackage{colortbl}
\usepackage{multicol}
\usepackage{booktabs}
\usepackage{makecell}
\usepackage{amsmath, amssymb}
\usepackage{multirow}
\usepackage{mathtools}
\usepackage{enumitem}
\usepackage{subcaption}
\usepackage{float}
\usepackage{graphicx}
\usepackage{caption} 
\usepackage{upgreek}
\usepackage{seqsplit}
\usepackage{color,soul}
\usepackage{arydshln}

\usepackage{amsmath,amsfonts,bm}

\def\eqref#1{equation~\ref{#1}}

\def\1{\bm{1}}

\def\vk{{\bm{k}}}

\def\vx{{\bm{x}}}
\def\vy{{\bm{y}}}

\DeclareMathAlphabet{\mathsfit}{\encodingdefault}{\sfdefault}{m}{sl}
\SetMathAlphabet{\mathsfit}{bold}{\encodingdefault}{\sfdefault}{bx}{n}

\usepackage{caption}
\usepackage{subcaption}
\definecolor{gg}{gray}{0.92}
\newcolumntype{a}{>{\columncolor{gg}}c}

\definecolor{lightgreen}{rgb}{0.56, 0.93, 0.56}

\DeclareRobustCommand{\hlpink}[1]{{\sethlcolor{pink}\hl{#1}}}
\DeclareRobustCommand{\hlyellow}[1]{{\sethlcolor{yellow}\hl{#1}}}
\DeclareRobustCommand{\hlgreen}[1]{{\sethlcolor{lightgreen}\hl{#1}}}

\title{Knowledge-Augmented Language Model Prompting \\ for Zero-Shot Knowledge Graph Question Answering}

\author{
    Jinheon Baek$^1$\thanks{\hspace{0.2cm} Work done while interning at Amazon. Corresponding author: Jinheon Baek (\texttt{jinheon.baek@kaist.ac.kr})} \quad
    \quad Alham Fikri Aji$^2$ \quad
    \quad Amir Saffari$^3$ \\
    KAIST$^{1}$ \quad MBZUAI$^{2}$ \quad Amazon$^{3}$ \\
    \texttt{jinheon.baek@kaist.ac.kr} \quad \texttt{alham.fikri@mbzuai.ac.ae} \quad \texttt{amsafari@amazon.com}
}

\begin{document}
\maketitle

\begin{abstract}

Large Language Models (LLMs) are capable of performing zero-shot closed-book question answering tasks, based on their internal knowledge stored in parameters during pre-training. However, such internalized knowledge might be insufficient and incorrect, which could lead LLMs to generate factually wrong answers. Furthermore, fine-tuning LLMs to update their knowledge is expensive. To this end, we propose to augment the knowledge directly in the input of LLMs. Specifically, we first retrieve the relevant facts to the input question from the knowledge graph based on semantic similarities between the question and its associated facts. After that, we prepend the retrieved facts to the input question in the form of the prompt, which is then forwarded to LLMs to generate the answer. Our framework, Knowledge-Augmented language model PromptING (KAPING), requires no model training, thus completely zero-shot. We validate the performance of our KAPING framework on the knowledge graph question answering task, that aims to answer the user's question based on facts over a knowledge graph, on which ours outperforms relevant zero-shot baselines by up to 48\% in average, across multiple LLMs of various sizes.

\end{abstract}
\section{Introduction}

\begin{figure}[t]

    \centering
    \includegraphics[width=0.99\columnwidth]{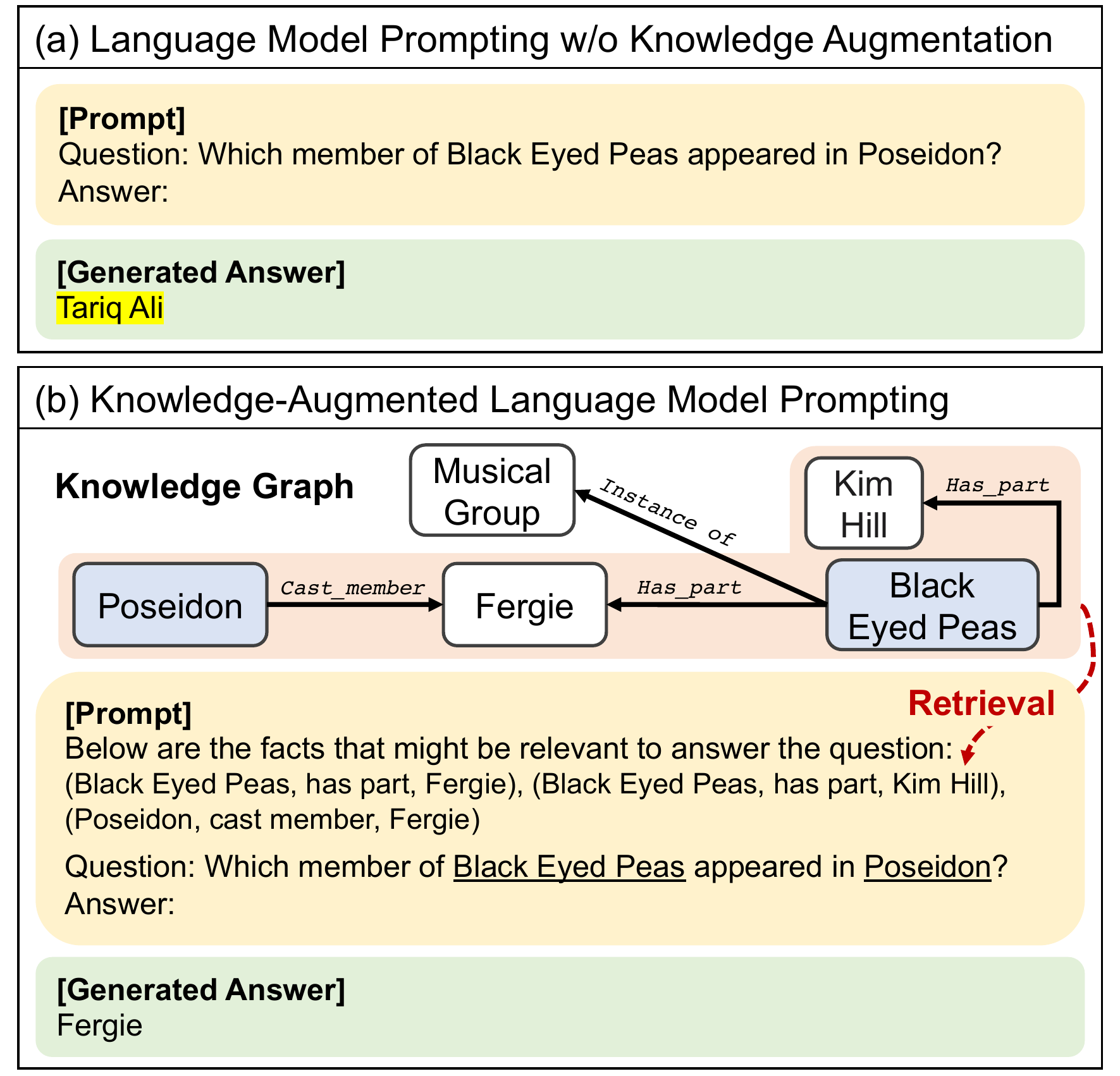}
    \vspace{-0.275in}
    \caption{(a) For the input question in the prompt, the large language model, GPT-3~\cite{gpt3}, can generate the answer based on its internal knowledge in parameters, but hallucinates it which is highlighted in yellow. (b) Our Knowledge-Augmented language model PrompTING (KAPING) framework first retrieves the relevant facts in the knowledge graph from the entities in the question, and then augments them to the prompt, to generate the factually correct answer.}
    \label{fig:concept}
    \vspace{-0.2in}
    
\end{figure}

Pre-trained Language Models (LMs)~\cite{bert, T5}, which are trained on a large amount of text corpora with self-supervised learning, can perform closed-book Question Answering (QA) tasks that aim to answer the user's question based only on their internal knowledge in parameters, without using any external knowledge~\cite{LMKB, LMKB/Finetune}. Also, when we increase the LM sizes, Large Language Models (LLMs) can generate the answer for the question without any additional fine-tuning steps, called \textit{LM prompting}~\cite{gpt3, PromptSurvey}. However, since the knowledge in LLMs might be incomplete, incorrect, and outdated, they often generate factually wrong answers, known as \textit{hallucination}~\cite{Hallucination} (See Figure~\ref{fig:concept}a). Also, refining the knowledge in LLMs with parameter updates is costly, especially when knowledge is constantly changing (e.g., exchange rates of money). Lastly, whether LLMs are fetching the correct knowledge for QA is unclear.

To overcome those limitations, we propose to retrieve and inject the relevant knowledge directly as an input, called a \textit{prompt}, to LLMs (Figure~\ref{fig:concept}b). As a knowledge source, we use a Knowledge Graph (KG) consisting of symbolic knowledge in the form of a triple: (head entity, relation, tail entity). Therefore, to extract the relevant facts to the input question, we first match entities in the question with entities in the KG. After that, triples associated to entities in the KG are verbalized (i.e., transforming the symbolic relational knowledge to the textual string) and prepended to the input question, which are then forwarded to LLMs to generate the answer. Consequently, LLMs conditioned on the factual knowledge are able to generate the factual answers, alleviating the hallucination issue, while keeping LLMs' parameters unchanged: fine-tuning is not required for knowledge updates. We refer to our overall framework as \textbf{K}nowledge-\textbf{A}ugmented language model \textbf{P}rompt\textbf{ING} (\textbf{KAPING}), which is completely \textit{zero-shot} and can be done with any off-the-shelf LLMs, without additional training.

While the above scheme looks simple yet effective, there is a couple of challenges. First, most retrieved triples associated with the question entities are unrelated to answer the given question. For example, when we retrieve the associated triples for the question entity (e.g., Poseidon) in Figure~\ref{fig:concept} in the Wikidata KG~\cite{wikidata}, there exist 60 triples, and most of them (e.g., genre, publication date, to name a few) are irrelevant to answer the question. Therefore, they might mislead the model into generating incorrect answers. On the other hand, the number of triples for the question entities is occasionally large (e.g., 27\% samples for the WebQSP dataset~\cite{WebQSP} have more than 1,000 triples), thereby encoding all triples including unnecessary ones yields high computational costs, especially on LLMs. 

To overcome such challenges, we further propose to filter out unnecessary triples based on their semantic similarities to the input question, inspired by the information retrieval~\cite{SemanticSearch}. To be specific, we first represent the question and its associated verbalized triples in the embedding space. Then, we retrieve the small number of triples whose embeddings are more close to the input question's embedding than others. By doing so, we can prepend only the more relevant triples to the given question, which can effectively prevent LLMs from generating irrelevant answers with high computational efficiencies, unlike the one that augments all triples. Note that, our filtering approach uses off-the-shelf sentence embedding models~\cite{MPNet, TAS-B}; thus no additional training is required in every part of our pipeline.

We then validate our KAPING framework on Knowledge Graph Question Answering (KGQA) tasks. The results show that our KAPING significantly outperforms relevant zero-shot baselines. Also, the detailed analyses support the importance of knowledge retrieval and augmentation schemes.

Our contributions in this work are threefold:
\vspace{-0.1in}
\begin{itemize}[itemsep=0.5mm, parsep=1pt]
  \item We present a new knowledge-augmented LM prompting framework that leverages the factual knowledge from KGs, for zero-shot QA.
  \item We propose to retrieve and augment relevant facts from KGs, based on semantic similarities between the question and its associated triples.
  \item We validate our KAPING on KGQA benchmark datasets, on which ours impressively outperforms relevant zero-shot baselines.
\end{itemize}

\vspace{-0.075in}
\section{Related Work}
\label{sec:related_work}

\paragraph{Language Model Prompting} 
Language model pre-training, which trains Transformers~\cite{transformer} on unannotated text corpora with auto-encoding~\cite{bert, roberta} or auto-regressive~\cite{XLNet, gpt} objectives, becomes an essential approach for natural language tasks. Also, Large Language Models (LLMs)~\cite{gpt3, T5, PaLM, AlexaTM} are able to perform zero-shot learning, for example, generating the answer for the input textual prompt, based on the knowledge stored in pre-trained parameters~\cite{LMKB, LMKB/Finetune, LMKB/Bio}, without additional parameter updates as well as labeled datasets. To further improve their performances, some work~\cite{prompt-retrieval/1, prompt-retrieval/2} proposes retrieving relevant samples to the input question from the training dataset and prepending them in the prompt under few-show learning. Recent few work~\cite{T0, FLAN} further shows that, when LLMs are fine-tuned on a collection of instructions phrased from natural language tasks, they can have strong generalization performance on unseen zero-shot tasks. However, the knowledge inside LMs might be insufficient to tackle factual questions, which gives rise to knowledge-augmented LMs. Notably, our LM prompting is different from prompt-tuning literature~\cite{prompt-tuning, prompt-tuning/retrieval} that additionally tunes LMs with model training (See Appendix~\ref{appendix:discussion:prompt} for discussions).

\paragraph{Knowledge-Augmented LMs} Recent work proposes to integrate the knowledge, such as documents from unstructured corpora (e.g., Wikipedia) and facts from Knowledge Graphs (KGs), into LMs. To mention a few, REALM~\cite{REALM} and RAG~\cite{RAG} learn to retrieve documents and augment LMs with them. In addition, KGs could be another knowledge source, where the knowledge is succinctly encoded in the most compact form, and some methods augment such facts in KGs into LMs~\cite{SE, DialoKG, SURGE}. However, all aforementioned approaches require massive amount of training data and model updates for downstream tasks. While more recent work~\cite{atlas} shows retrieval-augmented LM can have strong performance with few-shot learning, it still requires extra training steps, which is different from ours focusing on \textit{LM prompting} for entirely zero-shot. 

Recently, there are few studies augmenting the knowledge in the LM prompting scheme. At first, some work proposes to extract the knowledge in the parameters of LLMs themselves via prompting, and then use the extracted knowledge to answer the question~\cite{LMZeroShot, KnowledgePrompting, CoT/1, CoT/2}. However, since LLMs' parameters might be insufficient to store all the world knowledge, the extracted knowledge and generated answers might be inaccurate. On the other hand, most recently, \citet{WebAugment} propose to use the Google Search to retrieve documents on the Web, and then prepend the retrieved documents to the input question along with few-shot demonstrations, to answer the question under few-shot LLM prompting schemes. However, our focus on \textit{zero-shot prompting with KGs} is orthogonal to the previous study working on documents with few-shot prompting, and leveraging KGs can bring additional advantages. Specifically, since KGs can succinctly encode the knowledge in the compact triple form, for QA tasks, ours makes LLM prompting more efficient (i.e., reducing the input sequence length compared to the document case), as well as more effective on the zero-shot QA scheme: LLMs need to select one triple containing the answer entity in the prompt, instead of looking through lengthy documents having various entities. 

\paragraph{Knowledge Graph Question Answering}
The goal of our target Knowledge Graph Question Answering (KGQA) tasks is to answer the input question based on a set of facts over KGs~\cite{KBQA/1, KBQA/2}. Previous approaches are broadly classified into neural semantic parsing-based methods~\cite{SP/1, SP/2, SP/3}, information retrieval-based methods~\cite{IR/1, IR/2, IR/3}, and differentiable KG-based methods~\cite{Diff/1, Diff/2, Diff/3}, which, however, require annotated data with additional model training. While \citet{KGQA/Zeroshot} aim to transfer the KGQA model to the target language domains without any training data on them, this work indeed needs the labeled data to train the model on data-rich source domains first before transferring the model to the target domains. In contrast to all the aforementioned methods, we explore the novel zero-shot KGQA mechanism, which does not require any annotated QA pairs and additional training, leveraging LM prompting.

\section{Method}
\vspace{-0.02in}

We now describe our Knowledge-Augmented language model PromptING (KAPING) framework.

\subsection{LM Prompting for Zero-Shot QA}
\label{sec:preliminary}
\vspace{-0.02in}

We begin with the zero-shot question answering, and then explain the language model prompting.

\vspace{-0.03in}
\paragraph{Zero-Shot Question Answering} 
Given an input question $\vx$, the Question Answering (QA) system returns an answer $\vy$, where $\vx$ and $\vy$ consist of sequences of tokens: $\vx = [w_1, w_2, \ldots, w_{|\vx|}]$. Let $P$ be a QA model based on the generative Language Model (LM)~\cite{T5, gpt3}, which generates the conditional probability of answer $\vy$ for question $\vx$ as follows: $P(\vy | \vx)$. Then, in contrast to supervised learning that trains model $P$ with a set of annotated ($\vx$, $\vy$) pairs, zero-shot learning does not use any labeled samples and model training. Notably, we are interested in this zero-shot QA, since collecting the dataset and then fine-tuning the existing LMs for every new domain are known to be expensive and sometimes infeasible~\cite{zeroshot/justification/1, zeroshot/justification/2}.

\vspace{-0.03in}
\paragraph{LM Prompting} LMs are often pre-trained by predicting the next token based on previous tokens, which is known as auto-regressive language modeling~\cite{gpt, T5}. Then, thanks to this pre-training objective, LLMs can perform zero-shot instruction learning. Specifically, when we provide a question as well as an instruction (e.g., "Please answer the question: Who is the author of Lady Susan?") to the LLM (i.e., $P$), such the LLM, conditioned by the input text, can sequentially generate the probability of output tokens, which might be an answer, "Jane Austen". 

To be more formal, for every input question $\vx$, we first modify it with a particular instruction template $T$ into a textual string $\vx'$ called a \textit{prompt}, as follows: $T: \vx \mapsto \vx'$. For example, if we have the previous question $\vx = $ "Who is the author of Lady Susan?" along with the previous instruction template "Please answer the question:", the resulting prompt $\vx'$ would be $T(\vx) = $ "Please answer the question: Who is the author of Lady Susan?". Then, we forward the prompt $\vx'$ to the LLM (i.e., $P$), which then generates the answer (i.e., $\vy$) through $P(\vy | \vx')$. Note that this LM prompting scheme does not require any additional model parameter updates (i.e., fine-tuning) on the labeled data, thus appropriate for the target zero-shot QA task. 

However, there are multiple challenges in this naive zero-shot prompting for QA. First, LLMs, which rely on the knowledge in parameters, are vulnerable from generating the factually incorrect answer, since the knowledge in LLMs might be inaccurate, and outdated: knowledge can be emerged and changed over time. Also, refining the internalized knowledge with additional parameter updates is expensive, while it is necessary to reflect the wrong and ever growing knowledge. Lastly, which knowledge LLMs memorize and utilize when generating the answer to the question prompt is unclear, which limits their explainability on the outputs. 

\subsection{Knowledge-Augmented LM Prompting}
\label{sec:KAPING}

In order to tackle the aforementioned limitations of the existing LM prompting scheme, we propose to inject the relevant knowledge to the input question from the Knowledge Graph (KG), which we refer to as Knowledge-Augmented language model PromptING (KAPING). In this subsection, we first define the main objective of our KAPING framework, and then introduce the ingredients for augmenting the knowledge over KGs to LM prompts.

\paragraph{LM Prompting with Knowledge Graphs}
Instead of relying on the knowledge internalized in parameters, we propose to additionally access and inject the knowledge from the external KG, which contains accurate and up-to-date facts helpful to answer the question. Formally, a knowledge graph $\mathcal{G}$ consists of a set of factual triples $\left\{ (s, r, o) \right\}$, where $s$ and $o$ denote subject and object entities, and $r$ is a specific type of a relation between them. For example, one relational knowledge "Lady Susan was written by Jane Austen" can be represented as a triple consisting of two entities $s =$ "Lady Susan" and $o =$ "Jane Austen" along with a relation $r =$ "written by". Then, for the question prompt  $\vx'$ transformed from the example question $\vx = $ "Who is the author of Lady Susan?" via the template $T$, we additionally augment its relevant triple: (Lady Susan, written by, Jane Austen), to the LM prompting scheme. By doing so, LLMs can generate the correct answer with regard to the augmented knowledge from KGs, formalized as follows: $P(\vy | \vx', \mathcal{G})$. Note that, since we can provide specific and valid facts in KGs to LLMs whenever they exist, our framework can alleviate hallucination issue, originated from inaccurate and outdated knowledge in LLMs, without costly updating their model parameters. Furthermore, we can confirm whether LLMs generate answers based on augmented facts, thus improving the explainability of LM prompting. 

The remaining questions are then how to \textit{access} the relational symbolic facts over the KG from the input question, \textit{verbalize} the symbolic knowledge to the textual string, and \textit{inject} the verbalized knowledge into the LM prompting scheme. We explain them one by one in the following paragraphs.

\paragraph{Knowledge Access}
In order to utilize the related facts to the input question, we first extract the entities in the question. For example, for the question "Who is the author of \textit{Lady Susan}?", we extract the entity "Lady Susan". Then, based on the extracted entity, we find its corresponding entity over the KG, whose incident triples then become associated facts to the input question. Note that entity matching can be done by existing entity linking techniques~\cite{wu2019zero, li2020efficient, ReFinED}.

\paragraph{Knowledge Verbalization}
LLMs are working on textual inputs, whereas factual triples are represented over the symbolic graph. Therefore, before injecting the symbolic fact from KGs to LLMs, we first transform the triple consisting of $(s, r, o)$ into its textual string, called verbalization. While there exists recent methods~\cite{Verbalization/1, Verbalization/2} that particularly design or even learn the graph-to-text transformation, in this work, we use the linear verbalization: concatenating the subject, relation, and object texts in the triple, which we observe works well in LM prompting (See Appendix~\ref{appendix:results:verbalization}). For instance, one triple (Lady Susan, written by, Jane Austen) is used as is: "(Lady Susan, written by, Jane Austen)", for an LLM's input. 

\paragraph{Knowledge Injection}
Based on verbalized facts associated with the input question, the remaining step is to realize the knowledge injection mechanism, which allows LLMs to be grounded on the external knowledge, useful to generate the answer. Let assume we have a set of $N$ associated triples $\vk = \left\{ (s_i, r_i, o_i) \right\}_{i=1}^N$ for question $\vx$. Then, similar to instruction template $T: \vx \mapsto \vx'$ described in Section~\ref{sec:preliminary}, we modify $N$ verbalized triples $\vk$ along with the instruction for the knowledge injection into the knowledge prompt $\vk'$, as follows: $T: \vk \mapsto \vk'$. One particular template we use for constructing the prompt is that, we first enumerate $N$ verbalized triples line-by-line and then add the specific instruction: "Below are facts in the form of the triple meaningful to answer the question.", at the top of the prompt. After that, such the knowledge prompt string, $\vk'$, is prepended to the question prompt $\vx'$, and LLMs conditioned by knowledge and question prompts then sequentially generate the answer tokens, formalized as follows: $P(\vy | [\vk', \vx'])$, where $[\cdot]$ denotes concatenation.

\subsection{Question-Relevant Knowledge Retrieval}

The proposed KAPING framework in Section~\ref{sec:KAPING}, allows LLMs to leverage the knowledge from KGs for zero-shot QA. However, there are critical challenges that the number of triples associated to questions is often too large to forward in LLMs. Also, most of them are unrelated to the question, misleading LLMs into generating the irrelevant answer. 

\vspace{-0.05in}
\paragraph{Knowledge Retriever} To overcome those limitations, we further propose to retrieve and augment only the relevant triples to the question. Note that there exists a document-retrieval scheme~\cite{IR}, whose goal is to retrieve relevant documents for the given query based on their embedding similarities, which motivates us to retrieve, in our case, the triples for the user's question. In particular, thanks to the verbalizer defined in Section~\ref{sec:KAPING}, we can play with triples, obtained from symbolic KGs, over the text space. Therefore, for the verbalized triple and the question, we first embed them onto the representation space with off-the-shelf sentence embedding models for text retrieval~\cite{MPNet, DPR, ANCE}, and then calculate their similarities. After that, we use only the top-$K$ similar triples, instead of using all $N$ triples, associated to the given question. Note that, unlike few recent studies~\cite{Verbalization/1, Verbalization/2, SURGE} that aim at improving KG retrievers themselves under supervised training, we focus on zero-shot LM prompting with KGs, thus we use any off-the-shelf retrievers as a tool to filter out unnecessary triples for questions.

\begin{table*}[t!]
\caption{\textbf{Main results of language model prompting}, where we report the generation accuracy. The number inside the parentheses in the first row denotes the parameter size of language models, and best scores are emphasized in bold.}
\vspace{-0.1in}
\label{tab:main}
\small
\centering
\resizebox{\textwidth}{!}{
\begin{tabular}{llccccccccccca}
\toprule

\textbf{Datasets} & \textbf{Methods} & \textbf{T5 \scriptsize(0.8B)} & \textbf{T5 \scriptsize(3B)} & \textbf{T5 \scriptsize(11B)} & \textbf{OPT \scriptsize(2.7B)} & \textbf{OPT \scriptsize(6.7B)} & \textbf{OPT \scriptsize(13B)} & \textbf{T0 \scriptsize(3B)} & \textbf{T0 \scriptsize(11B)} & \textbf{GPT-3 \scriptsize(6.7B)} & \textbf{GPT-3 \scriptsize(175B)} & \textbf{AlexaTM \scriptsize(20B)} & \textbf{Average} \\

\midrule
\midrule

\multirowcell{5}[-0.5ex][l]{\textbf{WebQSP} \\ \textbf{w/ Freebase}} 

& No Knowledge & 6.95 & 13.40 & 9.48 & 19.85 & 29.77 & 28.38 & 21.43 & 40.77 & 44.63 & 63.59 & 46.79 & 29.55 \\

& Random Knowledge & 21.55 & 19.15 & 17.57 & 28.07 & 31.73 & 33.31 & 32.62 & 51.20 & 51.01 & 65.87 & 57.37 & 37.22 \\

& Popular Knowledge & 15.30 & 16.88 & 18.39 & 28.32 & 28.13 & 24.21 & 27.05 & 47.22 & 45.58 & 62.26 & 54.91 & 33.48 \\

& Generated Knowledge & 6.19 & 7.84 & 6.76 & 7.46 & 11.50 & 8.22 & 19.41 & 38.81 & 45.89 & 62.14 & 35.13 & 22.67 \\

\noalign{\vskip 0.25ex}\cdashline{2-14}\noalign{\vskip 0.75ex}

& \textbf{KAPING (Ours)} & \textbf{34.70} & \textbf{25.41} & \textbf{24.91} & \textbf{41.09} & \textbf{43.93} & \textbf{40.20} & \textbf{52.28} & \textbf{62.85} & \textbf{60.37} & \textbf{73.89} & \textbf{67.67} & \textbf{47.94} \\

\midrule

\multirowcell{5}[-0.5ex][l]{\textbf{WebQSP} \\ \textbf{w/ Wikidata}} 

& No Knowledge & 10.30 & 18.42 & 15.21 & 23.94 & 33.77 & 32.40 & 24.56 & 44.20 & 48.50 & 67.60 & 42.41 & 32.85 \\

& Random Knowledge & 17.94 & 22.78 & 24.28 & 37.24 & 35.61 & 38.27 & 28.85 & 47.68 & 52.05 & 60.64 & 55.63 & 38.27 \\

& Popular Knowledge & 15.35 & 20.80 & 20.74 & 30.83 & 30.01 & 27.83 & 24.83 & 48.02 & 47.41 & 63.37 & 53.92 & 34.83 \\

& Generated Knowledge & 11.94 & 13.30 & 12.28 & 11.26 & 17.53 & 14.19 & 22.92 & 41.34 & 48.77 & 65.89 & 31.16 & 26.42 \\

\noalign{\vskip 0.25ex}\cdashline{2-14}\noalign{\vskip 0.75ex}

& \textbf{KAPING (Ours)} & \textbf{23.67} & \textbf{40.38} & \textbf{35.47} & \textbf{49.52} & \textbf{53.34} & \textbf{51.57} & \textbf{49.86} & \textbf{58.73} & \textbf{60.44} & \textbf{69.58} & \textbf{65.04} & \textbf{50.69} \\

\midrule

\multirowcell{5}[-0.5ex][l]{\textbf{Mintaka} \\ \textbf{w/ Wikidata}} 

& No Knowledge & 11.23 & 14.25 & 17.06 & 19.76 & 27.19 & 26.83 & 14.75 & 23.74 & 34.65 & 56.33 & 41.97 & 26.16 \\

& Random Knowledge & 17.59 & 18.19 & 18.83 & 28.11 & 26.58 & 28.36 & 16.10 & 26.15 & 32.98 & 51.56 & 46.02 & 28.22 \\

& Popular Knowledge & 17.56 & 18.09 & 18.73 & 26.97 & 27.08 & 23.10 & 16.74 & 27.15 & 32.48 & 53.16 & 46.41 & 27.95 \\

& Generated Knowledge & 13.61 & 14.61 & 14.29 & 11.87 & 14.96 & 16.24 & 14.46 & 23.13 & 33.12 & 55.65 & 34.58 & 22.41 \\

\noalign{\vskip 0.25ex}\cdashline{2-14}\noalign{\vskip 0.75ex}

& \textbf{KAPING (Ours)} & \textbf{19.72} & \textbf{22.00} & \textbf{22.85} & \textbf{32.94} & \textbf{32.37} & \textbf{33.37} & \textbf{20.68} & \textbf{29.50} & \textbf{35.61} & \textbf{56.86} & \textbf{49.08} & \textbf{32.27} \\

\bottomrule

\end{tabular}
}
\end{table*}

\begin{figure*}[ht]
    \begin{minipage}{0.5\linewidth}
    \small
    \centering
    \vspace{-0.02in}
    \resizebox{0.99\textwidth}{!}{
    \renewcommand{\arraystretch}{0.85}
    \begin{tabular}{llcccccccc}
    \toprule
    
    & & \multicolumn{4}{c}{\bf 1-Hop Retrieval} & \multicolumn{4}{c}{\bf 2-Hop Retrieval} \\
    \cmidrule(l{2pt}r{2pt}){3-6} \cmidrule(l{2pt}r{2pt}){7-10}
    \textbf{Datasets} & \textbf{Retrievers} & \textbf{MRR} & \textbf{Top-1} & \textbf{Top-10} & \textbf{Top-30} & \textbf{MRR} & \textbf{Top-1} & \textbf{Top-10} & \textbf{Top-30} \\
    
    \midrule
    \midrule
    
    \multirowcell{3}[0ex][l]{\textbf{WebQSP} \\ \textbf{w/ Freebase}}
    
    & Random & 12.50 & 7.21 & 25.09 & 34.64 & 1.50 & 0.70 & 2.65 & 5.37 \\

    & Popular & 8.58 & 5.31 & 15.93 & 24.53 & 1.59 & 0.95 & 2.72 & 4.68 \\
    
    & MPNet & \textbf{47.27} & \textbf{40.27} & \textbf{60.56} & \textbf{64.48} & \textbf{41.64} & \textbf{33.12} & \textbf{58.47} & \textbf{65.23} \\
    
    \midrule
    
    \multirowcell{3}[0ex][l]{\textbf{WebQSP} \\ \textbf{w/ Wikidata}}
    
    & Random & 9.50 & 3.62 & 22.58 & 40.72 & 1.31 & 0.00 & 2.80 & 8.59 \\

    & Popular & 8.52 & 4.57 & 15.89 & 35.47 & 4.63 & 4.02 & 5.53 & 6.62 \\
    
    & MPNet & \textbf{43.46} & \textbf{33.36} & \textbf{64.39} & \textbf{70.67} & \textbf{40.42} & \textbf{30.56} & \textbf{62.62} & \textbf{71.56} \\
    
    \midrule
    
    \multirowcell{3}[0ex][l]{\textbf{Mintaka} \\ \textbf{w/ Wikidata}}
    
    & Random & 4.80 & 1.85 & 11.48 & 22.03 & 0.91 & 0.14 & 1.78 & 5.15 \\
    
    & Popular & 6.09 & 3.09 & 12.51 & 20.47 & 0.24 & 0.04 & 0.28 & 1.24 \\
    
    & MPNet & \textbf{13.01} & \textbf{7.50} & \textbf{25.44} & \textbf{35.43} & \textbf{13.00} & \textbf{6.82} & \textbf{26.65} & \textbf{40.01} \\
    
    \bottomrule
    
    \end{tabular}
    }
    \vspace{-0.08in}
    \captionof{table}{\textbf{Retriever results.} We compare random model, popular model, and MPNet~\cite{MPNet}, on 1- and 2-hop retrievals.} 
    \label{tab:retrieval}
    \end{minipage}
    \hfill
    \begin{minipage}{0.49\linewidth}
    \vspace{-0.02in}
        \centering
        \includegraphics[width=1\columnwidth]{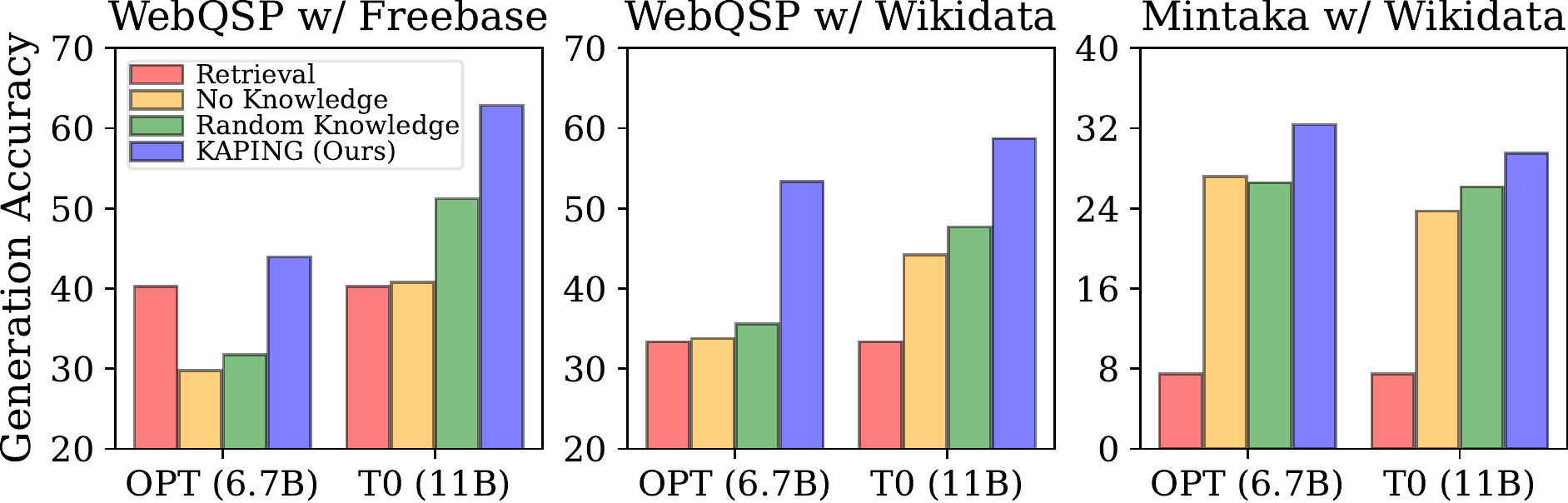}
    \vspace{-0.285in}
    \caption{\textbf{Comparisons of retrieval and LM prompting.} Retrieval is the Top-1 result of the MPNet~\cite{MPNet}. }
    \label{fig:retrieval}
    \end{minipage}
    \vspace{-0.175in}
\end{figure*}

\vspace{-0.03in}
\section{Experimental Setups}
\vspace{-0.075in}
We explain datasets, models, metrics, and implementations. For additional details, see Appendix~\ref{appendix:setups}.

\vspace{-0.03in}
\subsection{Datasets}
\vspace{-0.03in}
We evaluate our Knowledge-Augmented language model PromptING (KAPING) framework on two Knowledge Graph Question Answering (KGQA) datasets, namely WebQuestionsSP and Mintaka.

\vspace{-0.03in}
\paragraph{WebQuestionsSP (WebQSP)} This dataset~\cite{WebQuestions, WebQSP} is designed with a Freebase KG~\cite{Freebase}. It consists of 1,639 test samples, which we use for zero-shot evaluation. Additionally, since Freebase is outdated, we further use the Wikidata KG~\cite{wikidata} by using available mappings from Freebase ids to Wikidata~\cite{fbtowiki}. This additional dataset consists of 1,466 samples.

\vspace{-0.03in}
\paragraph{Mintaka} This dataset~\cite{Mintaka} is recently designed with the Wikidata KG for complex KGQA tasks. Among 8 different languages, we use English test sets consisting of 4,000 samples.

\subsection{Large Language Models}
\vspace{-0.03in}

To verify the performance of our KAPING framework on Large Language Models (LLMs), as well as benchmarking them on zero-shot KGQA, we use various LLMs with different sizes. Specifically, we use T5~\cite{T5} (0.8B, 3B, 11B), T0~\cite{T0} (3B, 11B), OPT~\cite{OPT} (2.7B, 6.7B) and GPT-3~\cite{gpt3} (6.7B, 175B). We provide details in Appendix~\ref{appendix:setups:llm}.

\subsection{Baselines and Our Model}
\vspace{-0.03in}

In this subsection, we explain four zero-shot LM prompting baselines and our KAPING framework.

\vspace{-0.05in}
\paragraph{No Knowledge} This is a naive LM prompting baseline, which generates answers from input questions without knowledge augmentation from KGs.

\vspace{-0.05in}
\paragraph{Random Knowledge} This is an LM prompting baseline, which additionally augments the randomly sampled $K$ triples, associated to the entities appeared in the question, to the prompt.

\vspace{-0.05in}
\paragraph{Popular Knowledge} This is an LM prompting baseline, which augments $K$ popular triples among all triples from the question entities, based on relations that appear the most frequently in the KG.

\vspace{-0.05in}
\paragraph{Generated Knowledge} This is an LM prompting baseline, which first extracts the knowledge from LLMs themselves based on prompting, and then augments them as the form of the prompt~\cite{KnowledgePrompting}, which is similar to~\citet{LMZeroShot}.

\vspace{-0.05in}
\paragraph{KAPING (Ours)} This is our Knowledge Augmented language model PromptING (KAPING) framework, which first retrieves the top-$K$ similar triples to the question with the knowledge retriever, and then augments them as the form of the prompt.

\vspace{-0.05in}
\subsection{Evaluation Metrics}
\label{sec:metrics}

\vspace{-0.05in}
\paragraph{Generation} Following the evaluation protocol of generative KGQA~\cite{GenerativeQA, Mintaka, KGQA/eval/partialmatch}, we use accuracy, which measures whether the generated tokens from the given prompt include one of the answer entities. Note that we further consider \textit{aliases} -- a set of alternative names -- of answer entities available in Freebase and Wikidata KGs, for evaluation.

\vspace{-0.05in}
\paragraph{Retrieval} We also measure the retriever performance, to see how much the retrieved triples are helpful for answer generation. As metrics, we use Mean Reciprocal Rank (MRR) and Top-K accuracy (Top-K), which are calculated by ranks of correctly retrieved triples containing answer entities among all triples associated to question entities.

\vspace{-0.05in}
\subsection{Implementation Details}
\label{sec:implementation_details}
\vspace{-0.05in}

For the knowledge injection, we set the number of retrieved facts as 10 ($K=10$), and the hop for triple retrieval as one. For the text-based retriever, we experiment with MPNet~\cite{MPNet} that uses the same encoder for embedding question and triples. See Appendix~\ref{appendix:setups:implementation} for additional details.

\vspace{-0.05in}
\section{Experimental Results and Analyses}
\label{sec:exp}
\vspace{-0.075in}

We provide the overall results of our KAPING framework along with its comprehensive analyses.

\paragraph{Main Results} As shown in Table~\ref{tab:main}, our KAPING framework significantly outperforms all LM prompting baselines, on zero-shot KGQA tasks. In particular, the generated knowledge model mostly degenerates the performance compared to the no knowledge model, since the extracted knowledge from LLMs themselves might be inaccurate. On the other hand, the random and popular knowledge baselines bring performance improvements, since the augmented knowledge from KGs are sometimes useful to answer the question. However, ours outperforms them, which suggests that, for zero-shot LM prompting for QA, the knowledge internalized in LLMs is insufficient to generate factual answers, and it is important to use only the relevant facts.

In addition, we also observe larger performance improvements when LMs are relatively small. In other words, since smaller models have insufficient parameter spaces to memorize the knowledge during pre-training, they are more likely to generate factually incorrect answers. However, when the appropriate knowledge is given to them, their performances sometimes become similar to larger models (e.g., different sizes of OPT have similar performances by our KAPING). Therefore, for tasks that require factual knowledge under low-resource setups (e.g., production), augmenting the knowledge would be beneficial, instead of increasing model sizes to handle the huge volume of knowledge.

\paragraph{Retriever Results}
To see how relevant the augmented knowledge is, we further measure the retrieval performances. As shown in Table~\ref{tab:retrieval}, the existing retrieval model (i.e., MPNet) shows superior performances against naive models: random and popular retrievers. This result suggests that our simple graph-to-text verbalization works well with the existing retriever, which further confirms that our KAPING augments useful facts in the LM prompt. Regarding the number of hops for the candidate triples to retrieve, we observe that, when we increase the hop-size from one to two, the retriever is more likely to retrieve irrelevant triples that does not include answer entities, as shown in Table~\ref{tab:retrieval}. Therefore, in our experiments, we retrieve knowledge among 1-hop triples of question entities.

\begin{figure}[t]
    \centering
    \caption{\textbf{Comparisons of correct and incorrect retrieval} for the generation performance on the GPT-3 (6.7B) model.}
    \label{fig:correct_retrieval}
    \vspace{-0.1in}
    \includegraphics[width=1\columnwidth]{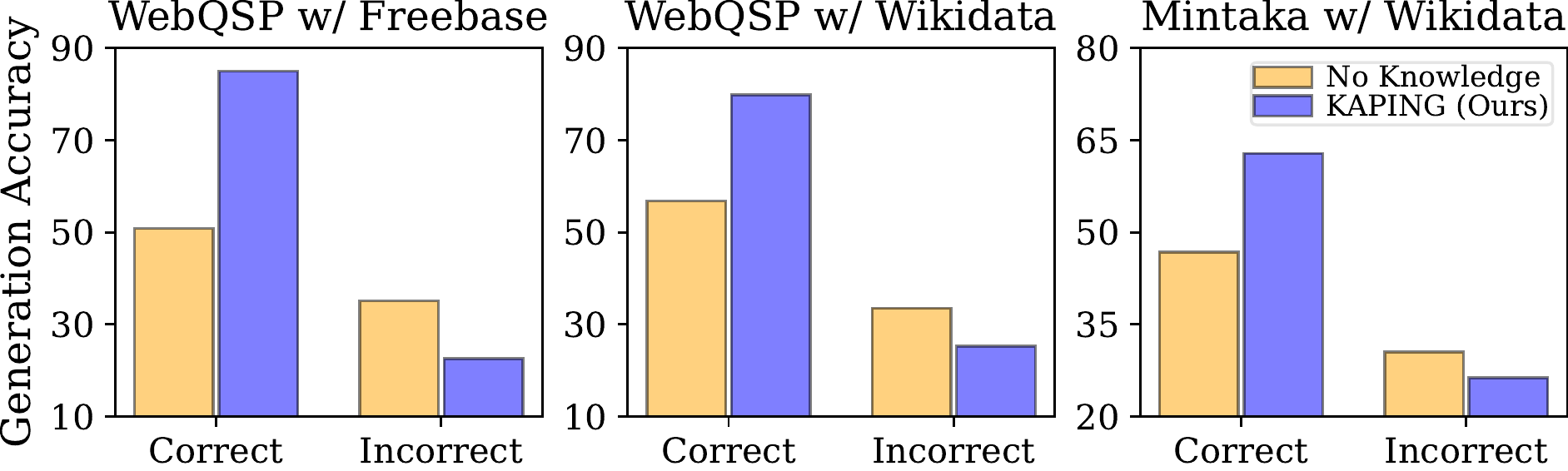}
\end{figure}
\begin{figure}[t]
    \vspace{-0.1in}
    \centering
    \includegraphics[width=1\columnwidth]{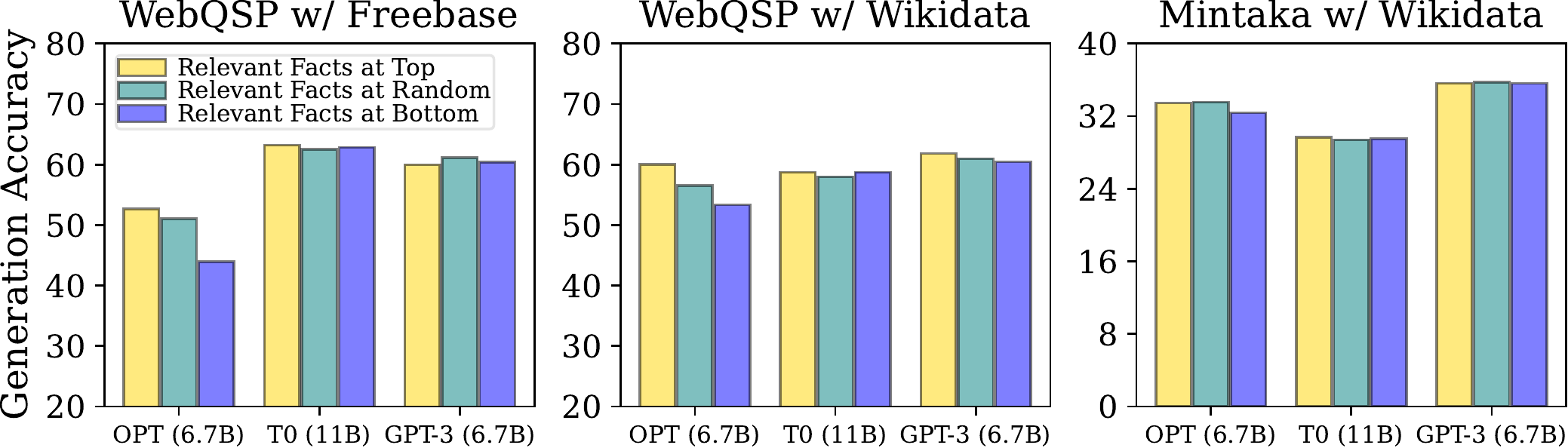}
    \vspace{-0.275in}
    \caption{\textbf{Performances with varying the knowledge order}, where we change the location -- top, bottom, or random -- of more relevant triples for the question in the prompt of LLMs.}
    \label{fig:knowledge_order}
    \vspace{-0.1in}
\end{figure}

\begin{figure}[t]

    \centering
    \caption{\textbf{Performances with varying knowledge amount}, where we change the number of retrieved triples to augment.}
    \label{fig:knowledge_size}
    \vspace{-0.09in}
    \includegraphics[width=1\columnwidth]{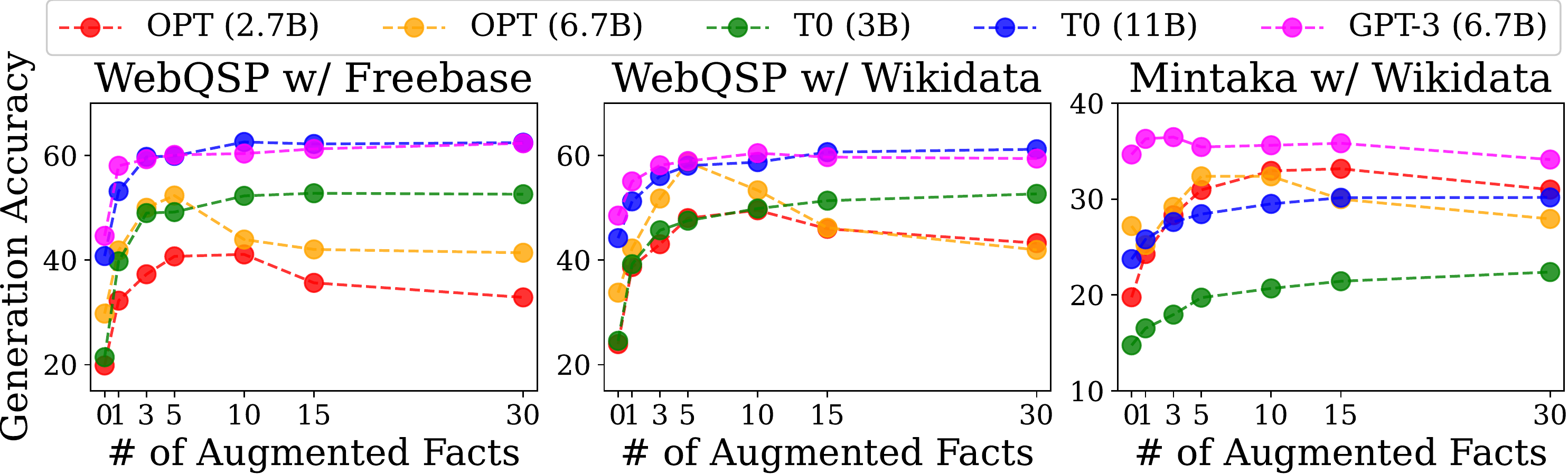}
    
\end{figure}
\begin{table}[t]
    \vspace{-0.075in}
    \centering
    \begin{center}
    \resizebox{0.47\textwidth}{!}{
    \renewcommand{\arraystretch}{0.53}
    \renewcommand{\tabcolsep}{5.0mm}
    \begin{tabular}[h]{lccc}
    \toprule
    & & \multicolumn{2}{c}{\bf Relative Time} \\
    \cmidrule(l{2pt}r{2pt}){3-4}
    \textbf{Models} & \textbf{\# of Retrieved Facts} & \textbf{T0 (3B)} & \textbf{OPT (2.7B)} \\
    \midrule
    No Knowledge & 0 & 1.00 & 1.00 \\
    \noalign{\vskip 0.25ex}\cdashline{1-4}\noalign{\vskip 0.75ex}
    \multirow{5}*{KAPING (Ours)} & 1 & 0.49 & 1.12 \\
    & 5 & 0.73 & 1.48 \\
    & 10 & 1.07 & 1.89 \\
    & 15 & 1.54 & 2.36 \\
    & 30 & 2.49 & 3.77 \\
    \bottomrule
    \end{tabular}
    }
    \end{center}
    \vspace{-0.125in}
    \caption{\small \textbf{Efficiencies with varying the knowledge amount}, where we measure the wall-clock time of every model for generating the answer on the WebQSP w/ Wikidata dataset.}
    \label{tab:efficiency}
    \vspace{-0.1in}
\end{table}

Additionally, since we can alternatively answer the input question based on entities in the Top-1 triple from the retriever, we compare the generation performance of LLMs to the retrieval performance. As shown in Figure~\ref{fig:retrieval}, LM prompting schemes even without knowledge augmentation (i.e., no knowledge) are superior than simply answering with the entity in the retrieved triple, except for the WebQSP w/ Freebase dataset. Also, we observe huge gaps between our KAPING framework and the simple retrieval scheme on all datasets. These results suggest that, for zero-shot KGQA, it would be helpful to leverage LLMs to generate answers based on their internalized and external facts, instead of directly searching answer entities over KGs. 

\setstcolor{red}

\begin{table*}[t!]\
\small
\centering
\caption{\small \textbf{Generation examples} of the prompted GPT-3 for the input question with augmented triples from the retriever, where, in the last row, we change the knowledge of augmented facts to see whether the model is able to adapt to the changed knowledge.}
\resizebox{1.00\textwidth}{!}{
\renewcommand{\arraystretch}{0.9}
\begin{tabular}{lll}

\toprule
\multicolumn{3}{p{\textwidth}}{\textbf{Question}: Where did Alex Chilton die?} \\

\multicolumn{3}{p{\textwidth}}{\textbf{Retrieved triples}: (Alex Chilton, place of death, \hlyellow{New Orleans}), (Alex Chilton, manner of death, natural causes), (Alex Chilton, cause of death, \hlpink{myocardial infarction}), (Alex Chilton, date of death, time: \hlgreen{+2010-03-17}), ...} \\

\multicolumn{3}{p{\textwidth}}{\textbf{Answer}: \hlyellow{New Orleans}} \\

\midrule
{} & \multicolumn{1}{c}{\textbf{Prompts}} & \multicolumn{1}{c}{\textbf{Outputs}} \\
\midrule

\textbf{No Knowledge}
& \multicolumn{1}{p{0.54\textwidth}}{Question: Where did Alex Chilton die? Answer:}  
&  \multicolumn{1}{p{0.45\textwidth}}{Alex Chilton died in a \st{car accident} on \st{February 10, 1979}.} \\

\noalign{\vskip 0.5ex}\cdashline{0-2}\noalign{\vskip 0.75ex}

\multirow{6}{*}{\textbf{KAPING (Ours)}} 
& \multicolumn{1}{p{0.54\textwidth}}{Below are facts in the form of the triple meaningful to answer the question. ... (Alex Chilton, date of death, time: \hlgreen{+2010-03-17}) (Alex Chilton, cause of death, \hlpink{myocardial infarction}) (Alex Chilton, manner of death, natural causes) (Alex Chilton, place of death, \hlyellow{New Orleans})
Question: Where did Alex Chilton die? Answer:}  
&  \multicolumn{1}{p{0.45\textwidth}}{Alex Chilton died on \hlgreen{March 17, 2010} in \hlyellow{New Orleans}, Louisiana due to a \hlpink{myocardial infarction}.} \\

\noalign{\vskip 0.5ex}\cdashline{0-2}\noalign{\vskip 0.75ex}

\multirowcell{6}[-0.0ex][l]{\textbf{KAPING (Ours)} \\ \textbf{w/ Knowledge Change}} 
& \multicolumn{1}{p{0.54\textwidth}}{Below are facts in the form of the triple meaningful to answer the question. ... (Alex Chilton, date of death, time: \hlgreen{+2000-09-01}) (Alex Chilton, cause of death, \hlpink{pancreatic cancer}) (Alex Chilton, manner of death, natural causes) (Alex Chilton, place of death, \hlyellow{Los Angeles})
Question: Where did Alex Chilton die? Answer:}  
&  \multicolumn{1}{p{0.45\textwidth}}{Alex Chilton died in \hlyellow{Los Angeles}, California on \hlgreen{September 1, 2000} from \hlpink{pancreatic cancer}.} \\

\bottomrule
\end{tabular}
}
\label{tab:case_study}
\vspace{-0.125in}
\end{table*}

\paragraph{Impact of Correct \& Incorrect Retrievals} 
We conduct analyses on how much the correctly retrieved triples, having answer entities, bring performance improvements, and how performances are affected by the incorrectly retrieved triples, which do not include answer entities. As shown in Figure~\ref{fig:correct_retrieval}, when retrieved triples contain answer entities, performances of LLMs are significantly improved, compared to models without knowledge augmentation. However, when retrievers fail, performances are lower than models of no knowledge augmentation. These results suggest, when relevant knowledge is augmented, LLMs can contextualize and generate answers accurately. Meanwhile, incorrectly retrieved knowledge makes LLMs condition on irrelevant facts, and generate wrong answers.

\paragraph{Varying the Amount of Knowledge} 
We change the number of facts, to see which triple amounts are optimal to augment in the prompt, by comparing trade-off between the generation performance and the wall-clock time. First of all, as shown in Figure~\ref{fig:knowledge_size}, most LLMs reach the somewhat highest performance, when the number of triples is 5 or 10. Also, when we further increase the augmented triple size to 15 and 30, performances of OPT models are largely decreasing. This result suggests that some LMs might be distracted by irrelevant triples when their volumes are high, therefore, failing to select and generate the answer entity.

We then measure the wall-clock time of the answer generation, for the encoder-decoder (T0) and decoder-only (OPT) models with varying the number of augmented triples in the prompt. As shown in Table~\ref{tab:efficiency}, regarding the encoder-decoder model, our KAPING framework with less than 10 triples is faster than the model without knowledge augmentation. We observe this is because, when the knowledge is augmented to the model, the model tends to generate shorter answers, which can reduce the decoding time. More specifically, the length of generated tokens for the T0 model with 10 triples is 15, whereas, the no knowledge model generates 32 tokens in average. However, for the decoder-only model (OPT), the more knowledge we augment, the slower the model becomes, because of its auto-regressive characteristic for digesting the input.

\begin{table}[t]
    \centering
    \begin{center}
    \resizebox{0.475\textwidth}{!}{
    \renewcommand{\arraystretch}{1.1}
    \begin{tabular}[h]{lccccccc}
    \toprule
    \textbf{Models} & \textbf{T5 \scriptsize(3B)} & \textbf{T5 \scriptsize(11B)} & \textbf{OPT \scriptsize(2.7B)} & \textbf{OPT \scriptsize(13B)} & \textbf{T0 \scriptsize(3B)} & \textbf{T0 \scriptsize(11B)} \\
    \midrule
    No Knowledge & 14.25 & 17.06 & 19.76 & 26.83 & 14.75 & 23.74 \\
    \noalign{\vskip 0.25ex}\cdashline{1-7}\noalign{\vskip 0.75ex}
    Random Knowledge & 18.19 & 18.83 & 28.11 & 28.36 & 16.10 & 26.15 \\
    Random Knowledge w/ EL & 15.99 & 17.98 & 23.10 & 26.47 & 15.60 & 24.66 \\
    \noalign{\vskip 0.25ex}\cdashline{1-7}\noalign{\vskip 0.75ex}
    KAPING & 22.00 & 22.85 & 32.94 & 33.37 & 20.68 & 29.50 \\
    KAPING w/ EL & 18.94 & 20.58 & 26.87 & 28.39 & 18.51 & 27.11 \\
    \bottomrule
    \end{tabular}
    }
    \end{center}
    \vspace{-0.15in}
    \caption{\small \textbf{Results with entity linking}, where the model w/ EL uses entities extracted from the entity linking technique~\cite{ReFinED}, instead of using labeled ones, on Mintaka.}
    \label{tab:er}
    \vspace{-0.125in}
\end{table}

\paragraph{Impact of Orders of Retrieved Triples} 
In few-shot LM prompting where LLMs additionally observe few examples in the prompt, they are known to be sensitive to the order of examples~\cite{prompt/order}, and they tend to follow the answer in the last example~\cite{regency/bias}. Based on those observations, we also conduct an analysis on whether the order of retrieved triples affects the performance. In particular, we vary the location of more similar triples for the question, by locating them at the Top, Bottom, or Random position of the prompt. As shown in Figure~\ref{fig:knowledge_order}, our KAPING is not sensitive to the location of retrieved triples, except for the OPT model on the WebQSP dataset. In other words, the OPT model tends to generate the entity located at the first part of the prompt input. Meanwhile, other LLMs can contextualize the entire prompt input, and generate the entity regardless of its position.

\paragraph{Effectiveness with Entity Linking} Following the conventional KGQA evaluation~\cite{Diff/1}, we use question entities labeled in datasets, to retrieve facts in KGs. However, to see the performance with entities identified by Entity Linking (EL) technique, we further conduct experiments with the EL model, namely ReFinED~\cite{ReFinED}. As shown in Table~\ref{tab:er}, while the performance of KAPING w/ EL is slightly decreasing from the model with labeled entities due to the performance of EL, we consistently observe meaningful performance improvements from a No Knowledge model.

\paragraph{Case Study}
We conduct a case study in Table~\ref{tab:case_study}. In particular, when the knowledge is not given to the LM, it hallucinates the factually incorrect answer. However, when related facts are retrieved and augmented in the prompt, it can generate the correct answer. In addition, we analyze whether our KAPING can adapt to the updated knowledge, motivated by that some knowledge can be changed over time, while the knowledge in LMs remains static. To do so, as shown in the last row of Table~\ref{tab:case_study}, we replace object entities of triples, and then forward the prompt with the modified facts to the LM. Then, the result shows that the LM can generate the output based on the updated facts, which suggests the potential of adapting LMs without costly updating their parameters.

\paragraph{Additional Results}
Note that we further provide additional experimental results in Appendix~\ref{appendix:results}. In particular, we compare the performance of retrievers in Appendix~\ref{appendix:results:retrieval}, conduct the sensitivity analysis on template texts in Appendix~\ref{appendix:results:template}, provide the results with additional metrics including human evaluation in Appendix~\ref{appendix:results:metrics}, validate our KAPING under few-shot setups in Appendix~\ref{appendix:results:fewshot}, provide the analysis on verbalization in Appendix~\ref{appendix:results:verbalization}, and provide the efficiencies in Appendix~\ref{appendix:results:efficiency}.
\section{Conclusion}
\vspace{-0.05in}
In this work, we focused on the limitation of existing LM prompting schemes, which rely on the static knowledge internalized in parameters; therefore, when such knowledge are incomplete, inaccurate, and outdated, LLMs may generate factually incorrect answers. To tackle this challenge, we introduced a novel Knowledge-Augmented language model PrompTING (KAPING) framework, which augments the knowledge for the input question from KGs directly in the input prompt of LLMs, with the fact retriever to inject only the relevant knowledge. The proposed framework is completely zero-shot, and versatile with any LMs, without additional parameter updates and training datasets. We validated that our KAPING yields huge performance gaps from the LM prompting model relying on its internal knowledge, especially with smaller LMs, on the KGQA tasks. We believe our new mechanism for augmenting facts from KGs to the LM prompt will bring substantial practical impacts in generating knowledge-grounded answers.

\section*{Limitations} 
\label{sec:limitation}
In this section, we faithfully discuss the current limitations and potential avenues for future research.

First of all, the generation performance of our knowledge-augmentation framework largely depends on the efficacy of retrievers. In other words, if the retriever fails to retrieve the relevant facts to the input question, the prompted LLM, conditioned on the irrelevant facts, is likely to generate the incorrect answer (See Figure~\ref{fig:correct_retrieval}). Similarly, if the retriever is not designed to retrieve the facts in 2-hop neighborhoods of the question entities, LLMs are less likely to generate the answer requiring 2-hop knowledge. Note that, for the Mintaka dataset~\cite{Mintaka}, the number of answerable questions with 1-hop facts is only 40\% of total samples. However, when we include 2-hop triples, the number of answerable questions becomes 62\%, which suggests the necessity of 2-hop retrievals, which is yet challenging (See Table~\ref{tab:retrieval}). Thus, future work may improve the retrieval scheme itself to provide more accurate facts including multi-hops to the LLM, or may develop the mechanism to prevent the LLM from being misled by unrelated facts. 

On the other hand, the evaluation metric for the generation performance of prompted LLMs may be further improved. Specifically, regarding our target KGQA tasks, the answer for the question is the entity in KGs. However, the prompted LLMs without additional training (i.e., zero-shot) tend to generate the answer as the sentence. For instance, the label entity for the question (e.g., Where did Alex Chilton die?) in Table~\ref{tab:case_study} is "New Orleans", however, the LLMs often generate the sentence-level output: "Alex Chilton died on March 17, 2010 in New Orleans, Louisiana due to a myocardial infarction". We currently evaluate the model performance by measuring whether generated tokens contain the answer entity or not; however, it would be worthwhile to develop the additional metric to compare the sentence-level output from LLMs to the word-level answer in KGs in a more effective way. Note that we also try other available metrics (See Appendix~\ref{appendix:results:metrics}), such as F1 and Exact Match (EM) scores~\cite{SQuAD}, however, they largely penalize the longer sentences (e.g., EM of correct examples in Table~\ref{tab:case_study} are 0), thus may not be appropriate for evaluating LM prompting schemes.

Lastly, since we focus on the improvement of knowledge injection in LM prompting, we use the labeled entities in KGQA datasets when evaluating models, following the existing KGQA evaluation setups~\cite{Diff/1, Diff/3}. However, in real-world applications where the entities in the question are mostly not provided, we first need to extract entities in the question with existing entity linking techniques; therefore, our model performance depends on the efficacy of entity linking. In particular, regarding the result with entity linking in Table~\ref{tab:er}, the portion of answerable questions from labeled entities in the dataset is 40\%, however, the portion of them with entities from the entity linking model~\cite{ReFinED} is 22\%. Therefore, since the improved entity linking performance would contribute to the performance gain of our KAPING framework, for KGQA tasks, future work may advance such the entity linking scheme.

\section*{Ethics Statement}
\vspace{-0.05in}
For a user's question, our knowledge-augmentation scheme can allow prompted LMs generate a factually correct answer, grounded by the provided knowledge, for KGQA tasks. However, the performance of our KAPING framework is still far from perfect, due to potential failures in entity linking, fact retrieval, and knowledge generation itself. Thus, we should be aware whether LMs generate correct answers, especially on high-risk domains.

\section*{Acknowledgements}
\vspace{-0.05in}
We thank the members of the End-to-End Reasoning team of Alexa AI at Amazon and the anonymous reviewers for their constructive comments. 

\bibliography{custom}
\bibliographystyle{acl_natbib}

\appendix

\clearpage

\section{Additional Experimental Setups}
\label{appendix:setups}
Here we provide additional experimental setups. 

\subsection{Datasets}
We provide the additional details for two Knowledge Graph Question Answering (KGQA) datasets, namely WebQuestionsSP and Mintaka, which we use for evaluating baselines and our model.

\paragraph{WebQuestionsSP (WebQSP)} A question and its corresponding answer are annotated with Freebase entities~\cite{Freebase}, and refined with additional cleaning steps~\cite{WebQSP}: filtering out samples with invalid annotations, from the original WebQuestions dataset~\cite{WebQuestions}.

\paragraph{Mintaka} This dataset~\cite{Mintaka} is designed for complex KGQA tasks including superlative and comparative questions, where question-answer pairs are collected from crowdsourcing with Wikidata entities~\cite{wikidata}.

\subsection{Large Language Models}
\label{appendix:setups:llm}
We describe the specific details of Large Language Models (LLMs) that we use for LM prompting.

\paragraph{T5} This model~\cite{T5} is an encoder-decoder model, and, among different variants, we use the LM-adapted version\footnote{https://github.com/google-research/text-to-text-transfer-transformer/blob/main/released\_checkpoints.md}, which is additionally pre-trained with auto-regressive language modeling objective~\cite{gpt} for LM prompting.

\paragraph{T0} This model~\cite{T0} is further fine-tuned from T5~\cite{T5} over prompted text-to-text tasks, for improved zero-shot generalization performance with LM prompting.

\paragraph{GPT-3} This model~\cite{gpt3} is a decoder only model, which we access via API\footnote{https://openai.com/api/}.

\paragraph{OPT} This model~\cite{OPT} is a decoder only model, freely available for researchers.

\paragraph{AlexaTM} This model~\cite{AlexaTM} is an encoder-decoder model, pre-trained with denoising, which reconstructs the context of 15\% dropped tokens, and auto-regressive, which predicts the next tokens based on their previous tokens, objectives.

\subsection{Evaluation Metrics}
\label{appendix:setups:metric}
We provide more details for evaluation metrics.

\paragraph{Aliases} For generative question answering tasks, there can be alternative names of entities, called aliases, and we consider them for evaluation. For example, one Wikidata entity, "William Shakespeare" (Q692), has alternative names, such as "Shakespeare" and "The Bard", and we consider them when measuring the generation performance.

\paragraph{Filtering Unnamed Entities} For evaluating generative models, the name of entities are required. However, we sometime cannot find the name of the answer entities from their ids on Freebase and Wikidata KGs. This is because the annotated answer entities are sometimes not entities but categories, and the entity ids in KGs could be changed but we cannot find the KG dumps that are used to annotate datasets. Therefore, we filter out samples that do not have literal name texts for the answer entities. This filtering step results in 1,582 test samples for the WebQSP w/ Freebase dataset, 1,466 test samples for the WebQSP w/ Wikidata dataset, and 2,814 test samples for the Mintaka dataset.

\subsection{Implementation Details}
\label{appendix:setups:implementation}
In this subsection, we provide additional details for implementing our KAPING framework.

\paragraph{Knowledge Injection Schemes} There are different choices in knowledge injection schemes, from the number of facts to retrieve, to the number of hops for candidate triples, to the order of retrieved facts in the prompt (i.e., where the most relevant knowledge should be located in the prompt), to the template of prompts including their instruction texts. While search spaces of them are extremely huge, we aim to to find the optimal one (See analyses in Section~\ref{sec:exp}). Specifically, as reported in Section~\ref{sec:implementation_details}, the best settings we find are the number of retrieved facts of 10, and the number of hops for the triples to retrieve from the question entities of one. Also, we locate more relevant triples to the input question closer to the question text in the prompt, inspired by the observation that the model tends to rewrite answers that appeared at the end of the prompt~\cite{regency/bias}. Further, we examine different instruction templates for generating answers, such as "Question: $\left\{ x \right\}$ Answer: " or "Please answer the following question: $\left\{ x \right\}$", where $x$ is the literal question. Regarding instruction templates, we observe that the performances of LLMs are sensitive across different instructions (See Appendix~\ref{appendix:results:template}), therefore, we try both of them and then report the best result.

\paragraph{Retrieval Models} To augment only the relevant triples to the input question under the zero-shot setup, we use off-the-shelf text-based retriever models. Specifically, we experiment with two different types of retrievers: symmetric retriever that uses the same encoder for question and triples; asymmetric one that uses individual encoders for them. For the symmetric retriever, we use MPNet~\cite{MPNet}, which is trained on 1B sentence pairs\footnote{https://huggingface.co/sentence-transformers/all-mpnet-base-v2}. Also, for the asymmetric retriever, we use TAS-B~\cite{TAS-B}, which is trained on the MSMARCO dataset~\cite{marco}. We mainly report the results with MPNet, unless noted, since there performances are similar (See Appendix~\ref{appendix:results:retrieval}).

\subsection{Hyperparameters and Resources}
\label{appendix:setups:hyp}

We evaluate all models with PyTorch~\cite{pytorch} and Transformers~\cite{wolf-etal-2020-transformers} libraries. We set the maximum number of input token lengths of LMs as 1,024 and the maximum number of output token lengths as 128, for encoder-decoder models. For decoder-only models, we set the maximum token lengths as 1,152 (1,024 + 128). For computing resources, we run all models with 8 V100 GPUs, having 8 $\times$ 32GB GPU memory, in which every model is runnable within one day. Note that, due to the expensive computational costs for model prompting with LLMs, we run every model one time, and then report the results, without additional hyperparameter tuning unless noted.

\section{Additional Experiment Results}
\label{appendix:results}
In this section, we provide additional experimental results, on the comparisons of available text-based retrieval models in Section~\ref{appendix:results:retrieval}, the sensitive analyses on template texts of the prompt in Section~\ref{appendix:results:template}, and the extra evaluation metrics in Section~\ref{appendix:results:metrics}.

\begin{table}[t]
    \centering
    \begin{center}
    \resizebox{0.475\textwidth}{!}{
    \renewcommand{\arraystretch}{1.1}
    \begin{tabular}{llcccccccc}
    \toprule
    
    & & \multicolumn{4}{c}{\bf 1-Hop Retrieval} & \multicolumn{4}{c}{\bf 2-Hop Retrieval} \\
    \cmidrule(l{2pt}r{2pt}){3-6} \cmidrule(l{2pt}r{2pt}){7-10}
    \textbf{Datasets} & \textbf{Retrievers} & \textbf{MRR} & \textbf{Top-1} & \textbf{Top-10} & \textbf{Top-30} & \textbf{MRR} & \textbf{Top-1} & \textbf{Top-10} & \textbf{Top-30} \\
    
    \midrule
    \midrule
    
    \multirowcell{2}[0ex][l]{\textbf{WebQSP} \\ \textbf{w/ Freebase}}

    & MPNet & 47.27 & 40.27 & 60.56 & 64.48 & 41.64 & 33.12 & 58.47 & 65.23 \\
    
    & TAS-B & 51.62 & 45.76 & 61.76 & 64.41 & 37.08 & 25.85 & 58.66 & 64.48 \\
    
    \midrule
    
    \multirowcell{2}[0ex][l]{\textbf{WebQSP} \\ \textbf{w/ Wikidata}}
    
    & MPNet & 43.46 & 33.36 & 64.39 & 70.67 & 40.42 & 30.56 & 62.62 & 71.56 \\
    
    & TAS-B & 46.68 & 37.65 & 65.08 & 70.67 & 41.92 & 32.20 & 62.21 & 72.17 \\
    
    \midrule
    
    \multirowcell{2}[0ex][l]{\textbf{Mintaka} \\ \textbf{w/ Wikidata}}
    
    & MPNet & 13.01 & 7.50 & 25.44 & 35.43 & 13.00 & 6.82 & 26.65 & 40.01 \\
    
    & TAS-B & 13.21 & 7.57 & 25.20 & 35.04 & 12.36 & 6.79 & 24.13 & 36.07 \\
    
    \bottomrule
    
    \end{tabular}
    }
    \end{center}
    \vspace{-0.15in}
    \caption{\small \textbf{Results of two different retrievers}, namely MPNet~\cite{MPNet} and TAS-B~\cite{TAS-B}.}
    \label{tab:additional:retrieval}
    \vspace{-0.15in}
\end{table}

\subsection{Performance Comparisons of Retrievers}
\label{appendix:results:retrieval}
In Table~\ref{tab:additional:retrieval}, we compare existing symmetric and asymmetric retrievers named MPNet~\cite{MPNet} and TAS-B~\cite{TAS-B}, explained in Section~\ref{appendix:setups:implementation}, on 1- and 2-hop retrievals. As shown in Table~\ref{tab:additional:retrieval}, we observe similar performances between symmetric (MPNet) and asymmetric (TAS-B) retrievers, which suggests that our simple graph-to-text verbalization is robust across different text-based retrieval schemes. Note that, since retrieval performances of both are similar, we conduct experiments mainly with MPNet, to reduce expensive computational costs for GPU usages.

\begin{table}[t]
    \centering
    \begin{center}
    \resizebox{0.475\textwidth}{!}{
    \renewcommand{\arraystretch}{1.1}
    \begin{tabular}{lllccccccc}
    \toprule
    
    \textbf{Datasets} & \textbf{Models} & \textbf{Templates} & \textbf{T5 \scriptsize(11B)} & \textbf{T0 \scriptsize(11B)} & \textbf{OPT \scriptsize(6.7B)} & \textbf{GPT-3 \scriptsize(6.7B)} \\
    
    \midrule
    \midrule
    
    \multirowcell{4}[0ex][l]{\textbf{WebQSP} \\ \textbf{w/ Freebase}}

    & \multirowcell{2}[0ex][l]{No Knowledge} & Default & 9.48 & 34.70 & 29.77 & 44.63 \\
    
    & & Please & 3.03 & 40.77 & 18.71 & 42.48 \\
    
    \noalign{\vskip 0.25ex}\cdashline{2-7}\noalign{\vskip 0.75ex}
    
    & \multirowcell{2}[0ex][l]{KAPING} & Default & 24.91 & 62.58 & 43.93 & 60.37 \\
    
    & & Please & 17.45 & 61.19 & 34.07 & 60.43\\
    
    \midrule
    
    \multirowcell{4}[0ex][l]{\textbf{WebQSP} \\ \textbf{w/ Wikidata}}

    & \multirowcell{2}[0ex][l]{No Knowledge} & Default & 15.21 & 38.88 & 33.77 & 48.50 \\
    
    & & Please & 5.12 & 44.20 & 22.71 & 48.29 \\
    
    \noalign{\vskip 0.25ex}\cdashline{2-7}\noalign{\vskip 0.75ex}
    
    & \multirowcell{2}[0ex][l]{KAPING} & Default & 35.47 & 58.73 & 53.34 & 60.44 \\
    
    & & Please & 20.12 & 56.89 & 48.16 & 59.69 \\
    
    \midrule
    
    \multirowcell{4}[0ex][l]{\textbf{Mintaka} \\ \textbf{w/ Wikidata}}

    & \multirowcell{2}[0ex][l]{No Knowledge} & Default & 17.06 & 22.60 & 27.19 & 35.00 \\
    
    & & Please & 5.47 & 23.74 & 17.70 & 34.65 \\
    
    \noalign{\vskip 0.25ex}\cdashline{2-7}\noalign{\vskip 0.75ex}
    
    & \multirowcell{2}[0ex][l]{KAPING} & Default & 22.85 & 29.50 & 32.37 & 33.55 \\
    
    & & Please & 14.68 & 29.18 & 28.18 & 35.61 \\
    
    \bottomrule
    
    \end{tabular}
    }
    \end{center}
    \vspace{-0.15in}
    \caption{\small \textbf{Results with varying instruction templates}, for various LLMs on the WebQSP and Mintaka datasets.}
    \label{tab:additional:template}
    \vspace{-0.15in}
\end{table}

\subsection{Sensitivity Analyses on Template Texts}
\label{appendix:results:template}
Following the observation in~\citet{regency/bias}, the performances of LLMs vary across different templates in the prompt. In our experiments, since it is computationally infeasible to try all different prompt templates on various LLMs, we consider two types of question templates, described in Appendix~\ref{appendix:setups:implementation}. In particular, for the question $x$, we use either "Question: $\left\{ x \right\}$ Answer: ", which we refer to as \textit{default} template, or "Please answer the following question: $\left\{ x \right\}$", referred to as \textit{please} template. As shown in Table~\ref{tab:additional:template}, for the T5 model, the default template is superior than the please template. Meanwhile, for the OPT model, the please template is superior than the other. However, for T0 and GPT-3 models, performance differences between default and please templates are marginal. Therefore, these results suggest that we may need to select instruction templates carefully across different LLMs for achieving optimal performances.

\begin{table*}[t!]
\small
\centering
\resizebox{\textwidth}{!}{
\renewcommand{\arraystretch}{0.95}
\begin{tabular}{llcccccccccccccccccc}
\toprule

& & \multicolumn{3}{c}{\textbf{T5 \scriptsize(0.8B)}} & \multicolumn{3}{c}{\textbf{T5 \scriptsize(3B)}} & \multicolumn{3}{c}{\textbf{T5 \scriptsize(11B)}} & \multicolumn{3}{c}{\textbf{OPT \scriptsize(2.7B)}} & \multicolumn{3}{c}{\textbf{OPT \scriptsize(6.7B)}} & \multicolumn{3}{c}{\textbf{OPT \scriptsize(13B)}} \\

\cmidrule(l{2pt}r{2pt}){3-5} \cmidrule(l{2pt}r{2pt}){6-8} \cmidrule(l{2pt}r{2pt}){9-11} \cmidrule(l{2pt}r{2pt}){12-14} \cmidrule(l{2pt}r{2pt}){15-17} \cmidrule(l{2pt}r{2pt}){18-20} 
\textbf{Datasets} & \textbf{Methods} & Acc. & F1 & EM & Acc. & F1 & EM & Acc. & F1 & EM & Acc. & F1 & EM & Acc. & F1 & EM & Acc. & F1 & EM \\

\midrule
\midrule

\multirowcell{5}[-0.5ex][l]{\textbf{WebQSP} \\ \textbf{w/ Freebase}} 

& No Knowledge & 6.95 & 5.20 & 0.00 & 13.40 & 8.11 & 0.00 & 9.48 & 8.25 & 0.06 & 19.85 & 7.20 & 0.38 & 29.77 & 10.60 & 0.06 & 28.38 & 7.92 & 0.70 \\

& Random Knowledge & 21.55 & 9.74 & 0.00 & 19.15 & 8.08 & 0.00 & 17.57 & 7.50 & 0.19 & 28.07 & 13.33 & 0.06 & 31.73 & 13.01 & 0.00 & 33.31 & 12.41 & 0.00 \\

& Popular Knowledge & 15.30 & 8.75 & 0.06 & 16.88 & 8.19 & 0.00 & 18.39 & 8.95 & 0.19 & 28.32 & 13.78 & 0.06 & 28.13 & 12.21 & 0.00 & 24.21 & 9.86 & 0.00 \\

& Generated Knowledge & 6.19 & 7.96 & 0.00 & 7.84 & 7.56 & 0.06 & 6.76 & 6.51 & 0.00 & 7.46 & 4.59 & 0.00 & 11.50 & 4.95 & 0.00 & 8.22 & 4.59 & 0.00 \\

\noalign{\vskip 0.25ex}\cdashline{2-20}\noalign{\vskip 0.75ex}

& \textbf{KAPING (Ours)} & 34.70 & 15.39 & 0.00 & 25.41 & 8.31 & 0.06 & 24.91 & 11.02 & 0.32 & 41.09 & 16.32 & 0.00 & 43.93 & 15.15 & 0.00 & 40.20 & 13.32 & 0.00 \\

\midrule

\multirowcell{5}[-0.5ex][l]{\textbf{WebQSP} \\ \textbf{w/ Wikidata}} 

& No Knowledge & 10.30 & 5.60 & 0.00 & 18.42 & 8.48 & 0.00 & 15.21 & 8.94 & 0.07 & 23.94 & 7.90 & 0.48 & 33.77 & 11.41 & 0.07 & 32.40 & 8.45 & 0.75 \\

& Random Knowledge & 17.94 & 7.81 & 0.00 & 22.78 & 7.74 & 0.07 & 24.28 & 9.41 & 0.34 & 37.24 & 16.78 & 0.00 & 35.61 & 12.54 & 0.00 & 38.27 & 14.61 & 0.07 \\

& Popular Knowledge & 15.35 & 8.01 & 0.00 & 20.80 & 8.48 & 0.00 & 20.74 & 9.20 & 0.14 & 30.83 & 15.65 & 0.00 & 30.01 & 13.32 & 0.00 & 27.83 & 11.95 & 0.00 \\

& Generated Knowledge & 11.94 & 8.64 & 0.00 & 13.30 & 8.19 & 0.07 & 12.28 & 7.11 & 0.00 & 11.26 & 5.06 & 0.00 & 17.53 & 5.60 & 0.00 & 14.19 & 4.94 & 0.00 \\

\noalign{\vskip 0.25ex}\cdashline{2-20}\noalign{\vskip 0.75ex}

& \textbf{KAPING (Ours)} & 23.67 & 10.46 & 0.00 & 40.38 & 13.25 & 0.00 & 35.47 & 11.50 & 0.34 & 49.52 & 20.17 & 0.00 & 53.34 & 16.62 & 0.00 & 51.57 & 16.73 & 0.14 \\

\midrule

\multirowcell{5}[-0.5ex][l]{\textbf{Mintaka} \\ \textbf{w/ Wikidata}} 

& No Knowledge & 11.23 & 6.77 & 0.00 & 14.25 & 9.81 & 0.00 & 17.06 & 10.28 & 0.00 & 19.76 & 6.63 & 0.28 & 27.19 & 10.60 & 0.04 & 26.83 & 9.82 & 0.43 \\

& Random Knowledge & 17.59 & 10.48 & 0.18 & 18.19 & 9.24 & 0.00 & 18.83 & 9.82 & 0.57 & 28.11 & 14.47 & 0.00 & 26.58 & 12.80 & 0.00 & 28.36 & 14.02 & 0.11 \\

& Popular Knowledge & 17.56 & 9.88 & 0.00 & 18.09 & 10.47 & 0.07 & 18.73 & 10.07 & 0.53 & 26.97 & 13.76 & 0.00 & 27.08 & 12.95 & 0.07 & 23.10 & 11.28 & 0.00 \\

& Generated Knowledge & 13.61 & 9.23 & 0.00 & 14.61 & 8.85 & 0.00 & 14.29 & 7.51 & 0.04 & 11.87 & 6.34 & 0.00 & 14.96 & 5.81 & 0.04 & 16.24 & 7.14 & 0.00 \\

\noalign{\vskip 0.25ex}\cdashline{2-20}\noalign{\vskip 0.75ex}

& \textbf{KAPING (Ours)} & 19.72 & 11.36 & 0.04 & 22.00 & 11.17 & 0.00 & 22.85 & 10.91 & 0.43 & 32.94 & 14.99 & 0.00 & 32.37 & 14.37 & 0.04 & 33.37 & 14.65 & 0.11 \\

\midrule
\midrule

& & \multicolumn{3}{c}{\textbf{T0 \scriptsize(3B)}} & \multicolumn{3}{c}{\textbf{T0 \scriptsize(11B)}} & \multicolumn{3}{c}{\textbf{AlexaTM \scriptsize(20B)}} & \multicolumn{3}{c}{\textbf{GPT-3 \scriptsize(6.7B)}} & \multicolumn{3}{c}{\textbf{GPT-3 \scriptsize(175B)}} & \multicolumn{3}{c}{\textbf{Average}} \\

\cmidrule(l{2pt}r{2pt}){3-5} \cmidrule(l{2pt}r{2pt}){6-8} \cmidrule(l{2pt}r{2pt}){9-11} \cmidrule(l{2pt}r{2pt}){12-14} \cmidrule(l{2pt}r{2pt}){15-17} \cmidrule(l{2pt}r{2pt}){18-20} 
\textbf{Datasets} & \textbf{Methods} & Acc. & F1 & EM & Acc. & F1 & EM & Acc. & F1 & EM & Acc. & F1 & EM & Acc. & F1 & EM & Acc. & F1 & EM \\

\midrule
\midrule

\multirowcell{5}[-0.5ex][l]{\textbf{WebQSP} \\ \textbf{w/ Freebase}} 

& No Knowledge & 21.43 & 22.70 & 9.99 & 40.77 & 46.10 & 34.39 & 46.79 & 17.65 & 0.00 & 44.63 & 21.12 & 1.77 & 63.59 & 32.75 & 8.47 & 29.55 & 17.05 & 5.07 \\

& Random Knowledge & 32.62 & 36.48 & 26.55 & 51.20 & 55.98 & 46.90 & 57.37 & 20.91 & 0.00 & 51.01 & 28.04 & 6.19 & 65.87 & 41.28 & 18.46 & 37.22 & 22.43 & 8.94 \\

& Popular Knowledge & 27.05 & 31.38 & 20.23 & 47.22 & 52.44 & 42.04 & 54.91 & 20.45 & 0.00 & 45.58 & 25.94 & 4.87 & 62.26 & 38.84 & 17.00 & 33.48 & 20.98 & 7.68 \\

& Generated Knowledge & 19.41 & 23.15 & 10.56 & 38.81 & 43.43 & 31.23 & 35.13 & 14.42 & 0.00 & 45.89 & 27.98 & 9.48 & 62.14 & 38.79 & 17.57 & 22.67 & 16.72 & 6.26 \\

\noalign{\vskip 0.25ex}\cdashline{2-20}\noalign{\vskip 0.75ex}

& \textbf{KAPING (Ours)} & 52.28 & 55.27 & 48.04 & 62.85 & 66.11 & 58.53 & 67.67 & 23.16 & 0.00 & 60.37 & 32.89 & 8.34 & 73.89 & 43.15 & 20.67 & \textbf{47.94} & \textbf{27.28} & \textbf{12.36} \\

\midrule

\multirowcell{5}[-0.5ex][l]{\textbf{WebQSP} \\ \textbf{w/ Wikidata}} 

& No Knowledge & 24.56 & 24.20 & 10.98 & 44.20 & 49.27 & 37.65 & 42.41 & 16.43 & 0.00 & 48.50 & 24.01 & 3.96 & 67.60 & 34.31 & 10.30 & 32.85 & 18.09 & 5.84 \\

& Random Knowledge & 28.85 & 33.08 & 22.37 & 47.68 & 52.34 & 42.50 & 55.63 & 19.88 & 0.06 & 52.05 & 25.37 & 2.18 & 60.64 & 36.88 & 13.92 & 38.27 & 21.49 & 7.41 \\

& Popular Knowledge & 24.83 & 27.89 & 16.03 & 48.02 & 52.84 & 41.88 & 53.92 & 19.77 & 0.00 & 47.41 & 24.36 & 3.75 & 63.37 & 37.08 & 14.73 & 34.83 & 20.78 & 6.96 \\

& Generated Knowledge & 22.92 & 25.28 & 11.80 & 41.34 & 45.70 & 33.83 & 31.16 & 13.36 & 0.00 & 48.77 & 29.72 & 11.19 & 65.89 & 39.52 & 17.87 & 26.42 & 17.56 & 6.80 \\

\noalign{\vskip 0.25ex}\cdashline{2-20}\noalign{\vskip 0.75ex}

& \textbf{KAPING (Ours)} & 49.86 & 50.75 & 41.27 & 58.73 & 61.90 & 53.27 & 65.04 & 22.72 & 0.00 & 60.44 & 31.18 & 6.82 & 69.58 & 41.83 & 19.71 & \textbf{50.69} & \textbf{27.01} & \textbf{11.05} \\

\midrule

\multirowcell{5}[-0.5ex][l]{\textbf{Mintaka} \\ \textbf{w/ Wikidata}} 

& No Knowledge & 14.75 & 20.84 & 11.34 & 23.74 & 28.69 & 20.86 & 41.97 & 17.05 & 0.00 & 34.65 & 17.67 & 2.31 & 56.33 & 26.77 & 6.11 & 26.16 & 14.99 & 3.76 \\

& Random Knowledge & 16.10 & 23.08 & 14.14 & 26.15 & 31.70 & 22.85 & 46.02 & 17.02 & 0.00 & 32.98 & 17.55 & 1.39 & 51.56 & 25.98 & 6.29 & 28.22 & 16.92 & 4.14 \\

& Popular Knowledge & 16.74 & 23.13 & 14.53 & 27.15 & 32.17 & 23.45 & 46.41 & 17.31 & 0.00 & 32.48 & 20.07 & 4.41 & 53.16 & 27.44 & 6.86 & 27.95 & 17.14 & 4.54 \\

& Generated Knowledge & 14.46 & 20.08 & 11.98 & 23.13 & 27.34 & 18.76 & 34.58 & 14.91 & 0.00 & 33.12 & 18.29 & 3.09 & 55.65 & 30.69 & 11.73 & 22.41 & 14.20 & 4.15 \\

\noalign{\vskip 0.25ex}\cdashline{2-20}\noalign{\vskip 0.75ex}

& \textbf{KAPING (Ours)} & 20.68 & 27.80 & 18.12 & 29.50 & 34.83 & 26.23 & 49.08 & 17.90 & 0.00 & 35.61 & 20.80 & 5.79 & 56.86 & 28.63 & 7.64 & \textbf{32.27} & \textbf{18.86} & \textbf{5.31} \\

\bottomrule

\end{tabular}
}
\vspace{-0.1in}
\caption{\textbf{LM prompting results with additional metrics: F1 and Exact Match (EM)}, along with accuracy (Acc.) scores.}
\vspace{-0.1in}
\label{tab:metric}
\end{table*}

Additionally, regarding the knowledge-injection template described in Section~\ref{sec:KAPING}, we also observe that the generation performance of GPT-3 depends on the instruction text in the template. In particular, we mainly conduct experiments with the template: "Below are facts in the form of the triple meaningful to answer the question."; however, we observe the performance degeneration when the augmented triples are irrelevant to the given question as shown in Figure~\ref{fig:correct_retrieval}. Therefore, to improve the performance on incorrect retrievals, we further experiment with the additional template: "Below are facts in the form of the triple that might be meaningful to answer the question.". Then, the GPT-3 (175B) model with the previous template achieves 74.16 and 42.80 accuracies for correct and incorrect retrievals, respectively. Meanwhile, the same model with the instruction template containing "might be" achieves 72.91 and 51.38 accuracies for correct and incorrect retrievals, respectively. Thus, these results suggest that the knowledge-injection template with "might be" statement makes the model less selective on the augmented triples while focusing more on the internalized knowledge in parameters, thus improving the incorrect retrieval performance while degenerating the correct retrieval.

\subsection{Additional Evaluation Metrics}
\label{appendix:results:metrics}
As described in Section~\ref{sec:metrics}, we evaluate the performance of LLMs based on whether generated tokens for the input question contain answer entities or not. This is because, as explained in Section~\ref{sec:limitation} of the limitation, pre-trained LLMs without further fine-tuning tend to generate the answer as the sentence, while the answer for the KGQA task is the entity consisting of few tokens. In this subsection, we further provide experiment results with additional evaluation metrics~\cite{SQuAD}, namely F1 and Exact Match (EM) scores. Note that they are frequently used for evaluating extractive QA models, whose goal is to classify the answer span in the given context, without generation. As shown in Table~\ref{tab:metric}, since the F1 score penalizes the longer sentence too much, the performances of LLMs evaluated by F1 scores are largely decreasing, except for the T0 model that is further fine-tuned by prompted text-to-text tasks, including QA, thus capable of generating entity-level outputs. Similarly, except for the T0, it is highly suboptimal to evaluate the performance of prompted LMs with EM scores, due to differences in output lengths. Thus, it would be promising direction to further develop better evaluation metrics for KGQA under LM prompting schemes, which we leave as future work. 

While such F1 and EM scores, used for extractive QA tasks, might be suboptimal to evaluate generative LM prompting schemes, our KAPING framework consistently outperforms all the other baselines based on averaged F1 and EM scores as well, by large margins. Note that the superior EM and F1 scores of the generated knowledge baseline with GPT-3 on few cases, even though they are rarely happen, is because, for this baseline, the GPT-3 model generates entity-level outputs, unlike ours that generates sentence-level outputs. In other words, the sentence-level outputs from our KAPING is often longer than the answer entities, since our model is grounded by retrieved facts from KGs as shown in Table~\ref{tab:examples_gpt3}; however, longer sentences penalize F1 and EM scores. More specifically, the average number of output sequence lengths of the generated knowledge model is 67.77, meanwhile, ours is 74.92. However, when we compare the generated knowledge baseline to our KAPING with other LLMs but also with other metrics, our KAPING significantly outperforms this baseline. 

\begin{table}[t]
    \centering
    \begin{center}
    \resizebox{0.475\textwidth}{!}{
    \renewcommand{\arraystretch}{1.05}
    \begin{tabular}[h]{llcccccc}
    \toprule
    \textbf{LLMs} & \textbf{Models} &\textbf{Correct} & \textbf{Semi-Correct} & \textbf{Incorrect} \\
    \midrule
    \multirowcell{2}[0ex][l]{\textbf{T0 (3B)}} 
    & No Knowledge & 7 & 1 & 22\\
    & KAPING (Ours) & 17 & 0 & 13 \\
    \midrule
    \multirowcell{2}[0ex][l]{\textbf{T0 (11B)}} 
    & No Knowledge & 14 & 0 & 16 \\
    & KAPING (Ours) & 20 & 0 & 10 \\
    \midrule
    \multirowcell{2}[0ex][l]{\textbf{GPT-3 (6.7B)}} 
    & No Knowledge & 12 & 4 & 14 \\
    & KAPING (Ours) & 19 & 4 & 17 \\
    \midrule
    \multirowcell{2}[0ex][l]{\textbf{GPT-3 (175B)}} 
    & No Knowledge & 22 & 1 & 7 \\
    & KAPING (Ours) & 26 & 1 & 3 \\
    \bottomrule
    \end{tabular}
    }
    \end{center}
    \vspace{-0.15in}
    \caption{\small \textbf{Human evaluation results}, where we randomly sample 30 examples from the WebQSP w/ Freebase dataset.}
    \label{tab:human}
    \vspace{-0.1in}
\end{table}

\paragraph{Human Evaluation} Additionally, similar to the previous generative QA work~\cite{LMKB/Finetune}, we manually inspect 30 samples from the WebQSP w/ Freebase dataset, to see whether the generated sentence is factually correct to the input question. For this experiment, we evaluate four LLMs: T0 (3B), T0 (11B), GPT-3 (6.7B), and GPT-3 (175B), with no knowledge baseline and our KAPING. Also, we use three different ratings for each generation example: 1) we label it as correct if all information in the generated sentence is factually correct to the question; 2) we label it as semi-correct if some information in the generated sentence is factually incorrect which yet contains at least one answer entity; 3) we label it as incorrect for all the other cases. As shown in Table~\ref{tab:human}, we observe that our KAPING framework can generate the factually correct answer more, compared to the no knowledge baseline, which are consistent with the results from available evaluation metrics in Table~\ref{tab:main} and Table~\ref{tab:metric}. We provide generated answers, which we use for human evaluation in Table~\ref{tab:human}, for GPT-3 (175B) and T0 (3B) models in Table~\ref{tab:examples_gpt3} and Table~\ref{tab:examples_t0}.

\begin{table}[t]
    \centering
    \begin{center}
    \resizebox{0.475\textwidth}{!}{
    \renewcommand{\arraystretch}{0.925}
    \begin{tabular}[h]{llcccccc}
    \toprule
    \textbf{Models} & \textbf{Shots} &\textbf{T5 \scriptsize(3B)} & \textbf{OPT \scriptsize(6.7B)} & \textbf{T0 \scriptsize(11B)} \\
    \midrule
    \multirowcell{3}[0ex][l]{No Knowledge} 
    & Zero-Shot & 18.42 & 33.77 & 44.20 \\
    & One-Shot & 18.28 & 36.90 & 41.13 \\
    & Three-Shots & 17.87 & 37.65 & 37.38 \\
    \midrule
    \multirowcell{3}[0ex][l]{KAPING (Ours)} 
    & Zero-Shot & 40.38 & 53.34 & 58.73 \\
    & One-Shot & 18.42 & 52.25 & 48.70 \\
    & Three-Shots & 10.16 & 50.34 & 43.45 \\
    \bottomrule
    \end{tabular}
    }
    \end{center}
    \vspace{-0.15in}
    \caption{\small \textbf{KGQA results with few-shot learning}. We vary the number of examples (i.e., shots) in the prompt, and report the performances on the WebQSP w/ Wikidata dataset.}
    \label{tab:fewshot}
    \vspace{-0.15in}
\end{table}

\subsection{Performances of Few-Shot Learning}
\label{appendix:results:fewshot}
While the focus of our work is zero-shot as outlined in the main paper, in this subsection, we additionally extend this zero-shot setting to the few-shot setting, where we prepend the few examples about the input-output pairs in the prompt of LLMs. As shown in Table~\ref{tab:fewshot}, for the KGQA task, the performances are decreasing when we increase the number of samples (i.e., shots) in the input prompt, except for the OPT model. We suggest this might be because, the injected examples in the prompt are less relevant to the given factual question, misleading the model to focus on unrelated contexts on the injected examples. This phenomenon is even more severe in our KAPING framework; this is similarly because our KAPING augments the retrieved facts, and if the facts on the other few-shot examples are further injected in the input prompt, the model is more likely to be confused by those irrelevant facts. For the OPT model, we observe a slight performance improvement in the No Knowledge model, since few injected examples provide a hint on how the output format looks like. We leave further extending our zero-shot KAPING framework to the few-shot learning mechanism as future work.

\begin{table}[t]
    \centering
    \begin{center}
    \resizebox{0.475\textwidth}{!}{
    \renewcommand{\arraystretch}{1.1}
    \begin{tabular}{llcccccccc}
    \toprule

    \textbf{Retrievers} & \textbf{MRR} & \textbf{Top-1} & \textbf{Top-10} & \textbf{Top-30} \\
    
    \midrule
    \midrule
    
    Random Retrieval & 9.50 & 3.62 & 22.58 & 40.72 \\

    Popular Retrieval & 8.52 & 4.57 & 15.89 & 35.47 \\

    Retrieval with Free-Form Texts & 41.33 & 31.11 & 62.07 & 69.92 \\

    Retrieval with Triple-Form Texts & 43.46 & 33.36 & 64.39 & 70.67 \\
    
    \bottomrule
    
    \end{tabular}
    }
    \end{center}
    \vspace{-0.15in}
    \caption{\small \textbf{Retrieval results with different verbalizers}. We use the graph-to-text transformation model proposed in~\citet{Verbalization/2} for obtaining free-form texts. For triple-form texts, we use the verbalization technique described in Section~\ref{sec:KAPING}. MPNet~\cite{MPNet} is used as the retriever, and the performance is reported on WebQSP w/ Wikidata.}
    \label{tab:additional:verbalization:retrieval}
    \vspace{-0.1in}
\end{table}
\begin{table}[t]
    \centering
    \begin{center}
    \resizebox{0.475\textwidth}{!}{
    \renewcommand{\arraystretch}{1.1}
    \begin{tabular}{llcccccccc}
    \toprule

    \textbf{Retrievers} & \textbf{T5 \scriptsize(3B)} & \textbf{OPT \scriptsize(6.7B)} & \textbf{T0 \scriptsize(3B)} & \textbf{T0 \scriptsize(11B)} \\
    
    \midrule
    \midrule

    No Knowledge & 18.42 & 33.77 & 24.56 & 44.20 \\

    KAPING with Free-Form Texts & 43.25 & 53.00 & 47.75 & 53.21 \\

    KAPING with Triple-Form Texts & 40.38 & 53.34 & 49.86 & 58.73 \\
    
    \bottomrule
    
    \end{tabular}
    }
    \end{center}
    \vspace{-0.15in}
    \caption{\small \textbf{KGQA results with different verbalizers}. We use the graph-to-text transformation model proposed in~\citet{Verbalization/2} for obtaining free-form texts. For triple-form texts, we use the verbalization technique described in Section~\ref{sec:KAPING}. We then inject the verbalized triples in the input prompt. We report the generation accuracy on WebQSP w/ Wikidata.}
    \label{tab:additional:verbalization:generation}
    \vspace{-0.18in}
\end{table}

\begin{table*}[t]
    \centering
    \begin{center}
    \resizebox{0.975\textwidth}{!}{
    \renewcommand{\arraystretch}{0.75}
    \renewcommand{\tabcolsep}{3.0mm}
    \begin{tabular}[h]{lccccccccc}
    \toprule
    & & \multicolumn{8}{c}{\bf Relative Time} \\
    \cmidrule(l{2pt}r{2pt}){3-10}
    \textbf{Models} & \textbf{\# of Augmented Knowledge} & \textbf{T5 (0.8B)} & \textbf{T5 (3B)} & \textbf{T5 (11B)} & \textbf{OPT (2.7B)} & \textbf{OPT (6.7B)} & \textbf{OPT (13B)} & \textbf{T0 (3B)} & \textbf{T0 (11B)} \\
    \midrule
    No Knowledge & 0 & 1.00 & 1.00 & 1.00 & 1.00 & 1.00 & 1.00 & 1.00 & 1.00 \\
    \noalign{\vskip 0.25ex}\cdashline{1-10}\noalign{\vskip 0.75ex}
    \multirow{5}*{\shortstack[l]{Document (Web) \\ Augmentation}} 
    & 1 & 1.20 & 1.45 & 2.13 & 1.43 & 1.65 & 1.63 & 1.60 & 2.29 \\
    & 5 & 2.78 & 4.16 & 6.80 & 3.42 & 3.90 & 3.66 & 2.98 & 9.01 \\
    & 10 & OOL & OOL & OOL & 6.44 & 7.36 & 6.67 & OOL & OOL \\
    & 15 & OOL & OOL & OOL & 9.35 & 10.71 & OOM & OOL & OOL \\
    & 30 & OOL & OOL & OOL & OOL & OOL & OOL & OOL & OOL \\
    \noalign{\vskip 0.25ex}\cdashline{1-10}\noalign{\vskip 0.75ex}
    \multirow{5}*{KAPING (Ours)} 
    & 1 & 1.08 & 0.97 & 1.35 & 1.12 & 1.21 & 1.19 & 0.49 & 1.28 \\
    & 5 & 1.22 & 1.50 & 2.13 & 1.48 & 1.65 & 1.60 & 0.73 & 2.18 \\
    & 10 & 1.53 & 2.10 & 3.11 & 1.89 & 2.20 & 2.10 & 1.07 & 3.83 \\
    & 15 & 1.84 & 2.74 & 4.02 & 2.36 & 2.76 & 2.58 & 1.54 & 4.59 \\
    & 30 & 2.82 & 4.42 & 6.05 & 3.77 & 4.28 & 4.06 & 2.49 & 7.76 \\
    \bottomrule
    \end{tabular}
    }
    \end{center}
    \vspace{-0.15in}
    \caption{\small \textbf{Efficiencies results}, where we measure the wall-clock time of every model for generating answers on the WebQSP w/ Wikidata dataset. The document augmentation model~\cite{WebAugment} augments documents listed in their paper, meanwhile, ours augments relevant triples to the question retrieved from KGs. We set the maximum number of input sequences for T5 and T0 models as 1,024, and for OPT as 2,048. OOL denotes the out-of-length errors, where the input prompt length exceeds the maximum input token lengths. OOM denotes the out-of-memory error on the machine having eight V100 GPUs.}
    \label{tab:appendix:efficiency}
    \vspace{-0.17in}
\end{table*}

\vspace{-0.025in}
\subsection{Analyses on Knowledge Verbalization}
\label{appendix:results:verbalization}
\vspace{-0.025in}

As described in the Knowledge Verbalization paragraph of Section~\ref{sec:KAPING}, we use the linear triple verbalization technique, which simply concatenates the tokens of subject, relation, and object in the triple, instead of using the sophisticated techniques that use the particular graph-to-text transformation methods~\cite{Verbalization/1, Verbalization/2}. This is because, we observe that our simple verbalization technique works well, and, in this subsection, we concretely show performance differences between our and existing verbalization techniques in both the knowledge retrieval and injection steps. Note that, for the comparison, we use the trained knowledge verbalizer proposed in~\citet{Verbalization/2}.

We first provide the fact retrieval performances across the different knowledge verbalization methods in Table~\ref{tab:additional:verbalization:retrieval}. As shown in Table~\ref{tab:additional:verbalization:retrieval}, we observe that our simple triple-form text verbalization is superior to the free-form text verbalization in the fact retrieval. This might be because the free-form verbalization model, transforming the graph to the text, might generate the incorrect output that is semantically different from the original triple, leading to the degenerated retrieval performances. 

On the other hand, we also report the generation results of KGQA with two different knowledge verbalizers on our KAPING framework in Table~\ref{tab:additional:verbalization:generation}. As shown in Table~\ref{tab:additional:verbalization:generation}, we observe that the performances between the free-form texts and the triple-form texts are comparable when augmented to LLMs with our KAPING framework. More specifically, for the T5 model, which is pre-trained on the unlabeled corpus without additional instruction tuning, the free-form text works well. Meanwhile, for the T0 model, which is further fine-tuned with natural language instruction tasks, it is beneficial to use our linear triple verbalizaton scheme.

\subsection{Additional Efficiency Comparisons}
\label{appendix:results:efficiency}
In this subsection, we further provide efficiency results of all LLMs that we use in our main experiments across three different models: no knowledge model, document augmentation (i.e., web augmentation) model~\cite{WebAugment}, and our KAPING framework. We note that, as discussed in the Knowledge-Augmented LMs paragraph of Section~\ref{sec:related_work}, the web augmentation method augments documents searched from Google with the few-shot learning setup. However, as we discuss there, this web augmentation is orthogonal to ours, since we use the completely different knowledge source (i.e., KGs) and our work is under the zero-shot learning setup; from which our core mechanisms of how to retrieve and augment relevant knowledge with LM prompting is clearly different and novel. Furthermore, as discussed in Section~\ref{sec:related_work}, this web augmentation method is infeasible to experimentally compare as well, since individual researches cannot freely access the Google Search API to retrieve documents for every question in the world. Also, it is computationally expensive to augment documents consisting of hundreds to thousands tokens~\cite{WebAugment} in LLMs, unlike our triple cases consisting of few tokens. In this subsection, to experimentally validate the latter issue, we further make the comparisons of computational costs between document augmentation and our fact augmentation. In particular, as shown in Table~\ref{tab:appendix:efficiency}, the answer generation speed of the web augmentation mechanism is significantly slower than our triple augmentation mechanism, since it requires more time to encode and condition documents in the input prompt compared to triples. Also, following the original paper~\cite{WebAugment}, the suggested number of documents to augment is 15, however, in the most cases, we observe out-of-length (OOL) errors, since the length of the input prompt with 15 documents is longer than the maximum input sequence length of LLMs. While our fact augmentation scheme is slower than the model without augmentation, we believe that, given the substantially improved performance in Table~\ref{tab:main} and the high efficiency compared to document augmentation in Table~\ref{tab:appendix:efficiency}, KAPING is highly beneficial.

\subsection{Result Analyses Across Question Types}
For the Mintaka dataset~\cite{Mintaka}, each question is belong to one of the following categories: Generic, Multihop, Intersection, Difference, Comparative, Superlative, Ordinal, Count, and Yes/No, which defines the complexity of questions. Therefore, to see which complexity category our knowledge-augmentation framework is helpful, and which category we should further improve on, we breakdown the performance of LLMs according to question types in Table~\ref{tab:appendix:complexity_type}. Note that, following the evaluation protocol in Section~\ref{appendix:setups:metric} where we filter out questions that do not have answer names, the Yes/No type questions are not considered. 

As shown in the last row of Table~\ref{tab:appendix:complexity_type} where we average the performance of all LLMs per category, our KAPING framework brings significant performance improvements on all categories except for the Comparative type. One particular comparative-type question is "Who has won more NBA Season MVPs, LeBron James or Steph Curry", and, since it is hard to retrieve and associate relevant triples for such the comparative-type question, our KAPING underperforms simple knowledge-injection baselines: random knowledge and popular knowledge. However, the KG-augmented models (e.g., random knowledge, popular knowledge, and our KAPING) outperform other baselines, which suggests that knowledge-augmentation mechanism is meaningful to tackle comparative questions, and one might further improve the retrieval scheme or the input prompt itself, which we leave as future work. 

On the other point we would like to mention is that, for the Count category, performances of T0 models are significantly low compared to other LLMs. This is surprising, since T0 models are further fine-tuned on the prompted text-to-text tasks, and they have strong performances on the other categories, thanks to fine-tuning. We believe such the low performance on the Count category is because, in the fine-tuning of T0 models, there are no prompted tasks related to counting, which makes T0 models hard to count particular instances. Therefore, to further improve the generalization performance of T0 models, one may additionally include more diverse prompted tasks, including the counting one, during the fine-tuning process.

\subsection{Generation Examples}

We provide generation examples for comparisons between the no knowledge baseline and our KAPING framework in Table~\ref{tab:examples_gpt3} and Table~\ref{tab:examples_t0} for GPT-3 and T0 language models, respectively. We also provide retrieved and generation examples of our KAPING framework with four different LLMs: T5 (11B), OPT (13B), T0 (11B), and GPT-3 (175B) on the WebQSP w/ Wikidata dataset in Table~\ref{tab:examples_ours}.

\section{Discussions on Prompt Design/Tuning}
\label{appendix:discussion:prompt}
We discuss differences between prompt design and prompt tuning, along with additional relevant work in the prompt tuning literature. As described in Section~\ref{sec:preliminary}, given an input question, the large language model can generate the answer text, which is called LM prompting~\cite{gpt3, PromptSurvey}. However, to further enhance the performance of models under the LM prompting scheme, prior work particularly designs the content in the prompt, which is called \textit{prompt design}~\cite{autoprompt, prompt/order}. More specifically, \citet{autoprompt} additionally include the particular trigger tokens, meaningful to the down-stream tasks, in the prompt, and \citet{prompt/order} change the order of demonstrations in the prompt under the few-shot LM prompting setup. Our method is in line with such the prompt design literature, and we introduce the method of knowledge augmentation in the input prompt with facts from KGs, to allow LLMs condition on factual knowledge for zero-shot QA. 

On the other hand, there exists \textit{prompt tuning} literature~\cite{prompt-tuning}, which additionally trains the prompt-relevant parameters with supervised learning objectives, while keeping the parameters of LLMs unchanged. While this prompt tuning approach can be beneficial in few-shot learning scenarios where the model is additionally tuned with few training examples, it is not suitable for our zero-shot learning. Also, unlike the prompt design approach, it is difficult to interpret and manipulate the prompt represented in the embedding space. 

Note that, recently, there are few knowledge-aware prompt tuning work~\cite{KnowPrompt, KPT, prompt-tuning/retrieval}, and, while they are fundamentally different from our LM prompting (i.e., prompt design), we additionally discuss them. First of all, \citet{KnowPrompt} tackle the relation extraction problem with prompt tuning, where they propose to embed the particular words related to the relation class in the embedding space. For example, for the relation type to classify: "county of birth", they embed person and country information in the representation space with training signals from supervised learning, for improved relation classification performance. Also, \citet{KPT} tackle the text classification task with prompt tuning, where they propose to not only consider the classification label word itself, but also the label word's related words. For example, for the sentence label "science", they further consider its related words: "physics" and "mathematics", defined in particular knowledge bases, such as WordNet~\cite{WordNet} and ConceptNet~\cite{ConceptNet}. Lastly, \citet{prompt-tuning/retrieval} tackle the similar text classification task with prompt tuning, where they propose to retrieve the data instance (i.e., a sentence and its label) in the training dataset based on the retriever training with supervised classification objectives. 

However, all the above knowledge-aware prompt tuning methods are clearly different from our proposed KAPING framework. At first, they are restricted to cloze-style prediction, in which they first include the particular mask token in the input prompt, and then classify the label (e.g., sentiment of the sentence, or relation in the given sentence) of the mask token, similar to the masked language modeling objective~\cite{bert, roberta}. Therefore, their cloze-style prediction schemes cannot be used for QA tasks, since the answer of the user's question is not the single token, and it is unclear to convert the predicted label token from the masked token to all different answers in the world. In contrast to them, our KAPING does not rely on the masked token classification scheme, thus ours is more flexible, and not restricted to cloze-style classification; suitable for answering any user's questions. Furthermore, some of them~\cite{prompt-tuning/retrieval, KnowPrompt} rely on training signals from the training dataset with supervised learning, meanwhile, ours is completely zero-shot. While \citet{prompt-tuning/retrieval} show the model's zero-shot ability, they require the training dataset as discussed in their paper, thus not suitable for our zero-shot QA as well. Lastly, we augment the factual knowledge by matching the entity in the question to its associated triples in KGs, however, prior work considers different knowledge source, which might not be helpful for QA tasks, such as relationships between words~\cite{KPT}, relationships between the relation class and particular words~\cite{KnowPrompt}, and a pair of sentence and its label in training data~\cite{prompt-tuning/retrieval}.

\begin{table*}[t]
    \centering
    \begin{center}
    \resizebox{0.99\textwidth}{!}{
    \renewcommand{\arraystretch}{1.1}
    \renewcommand{\tabcolsep}{1.5mm}
    \begin{tabular}[t]{llcccccccc}
    \toprule
    \textbf{LLMs} & \textbf{Models} & \textbf{Generic} {\scriptsize (557)} & \textbf{Multihop} {\scriptsize (220)} & \textbf{Intersection} {\scriptsize (396)} & \textbf{Difference} {\scriptsize (349)} & \textbf{Comparative} {\scriptsize (223)} & \textbf{Superlative} {\scriptsize (384)} & \textbf{Ordinal} {\scriptsize (307)} & \textbf{Count} {\scriptsize (378)} \\
    \midrule
    \multirow{5}*{T5 (0.8B)} 
    & No Knowledge & 7.00 & 3.64 & 8.08 & 7.45 & 69.06 & 2.86 & 2.61 & 10.05\\
    & Random Knowledge & 11.49 & 5.45 & 8.33 & 11.75 & 86.10 & 6.77 & 8.14 & 26.98 \\
    & Popular Knowledge & 13.82 & 5.91 & 11.62 & 8.60 & 87.00 & 8.33 & 5.86 & 22.22 \\
    & Generated Knowledge & 7.72 & 2.73 & 5.81 & 8.02 & 82.06 & 3.39 & 1.95 & 21.43 \\
    & KAPING (Ours) & 18.85 & 6.36 & 15.40 & 10.32 & 83.41 & 9.64 & 7.49 & 24.60 \\
    \midrule
    \multirow{5}*{T5 (3B)} 
    & No Knowledge & 10.41 & 4.09 & 9.60 & 9.74 & 71.30 & 5.47 & 4.56 & 17.99 \\
    & Random Knowledge & 17.41 & 6.82 & 13.64 & 14.61 & 55.16 & 8.59 & 7.82 & 30.42 \\
    & Popular Knowledge & 14.90 & 6.82 & 14.90 & 13.75 & 57.40 & 8.85 & 10.75 & 28.84 \\
    & Generated Knowledge & 7.90 & 3.64 & 8.33 & 8.31 & 82.51 & 4.69 & 3.91 & 21.96 \\
    & KAPING (Ours) & 25.31 & 12.27 & 20.96 & 15.76 & 47.98 & 10.68 & 9.77 & 35.71 \\
    \midrule
    \multirow{5}*{T5 (11B)} 
    & No Knowledge & 10.23 & 5.00 & 10.35 & 8.60 & 92.83 & 7.55 & 3.58 & 24.87 \\
    & Random Knowledge & 20.29 & 7.27 & 11.87 & 12.89 & 60.99 & 10.68 & 9.12 & 27.51 \\
    & Popular Knowledge & 16.88 & 7.27 & 12.88 & 13.18 & 72.20 & 9.11 & 10.42 & 24.34 \\
    & Generated Knowledge & 7.72 & 2.73 & 5.30 & 7.45 & 89.24 & 3.91 & 2.28 & 22.49 \\
    & KAPING (Ours) & 24.42 & 8.64 & 18.69 & 16.05 & 65.92 & 11.98 & 11.07 & 34.66 \\
    \midrule
    \multirow{5}*{OPT (2.7B)} 
    & No Knowledge & 24.06 & 10.00 & 16.67 & 10.32 & 54.26 & 20.05 & 14.98 & 14.29 \\
    & Random Knowledge & 29.44 & 13.18 & 23.74 & 18.34 & 93.27 & 15.62 & 14.01 & 34.13 \\
    & Popular Knowledge & 28.90 & 14.09 & 20.45 & 18.62 & 90.58 & 12.76 & 13.36 & 34.13 \\
    & Generated Knowledge & 7.90 & 6.82 & 10.35 & 8.02 & 44.84 & 4.19 & 4.56 & 20.11 \\
    & KAPING (Ours) & 33.75 & 15.91 & 34.85 & 20.63 & 93.27 & 15.89 & 19.54 & 43.65 \\
    \midrule
    \multirow{5}*{OPT (6.7B)} 
    & No Knowledge & 29.62 & 12.73 & 37.37 & 20.06 & 62.78 & 20.83 & 22.80 & 16.93 \\
    & Random Knowledge & 23.52 & 14.09 & 19.44 & 20.92 & 89.69 & 13.02 & 15.31 & 36.77 \\
    & Popular Knowledge & 24.42 & 13.18 & 24.24 & 22.92 & 83.86 & 14.84 & 17.26 & 32.80 \\
    & Generated Knowledge & 11.67 & 8.64 & 16.92 & 12.61 & 43.95 & 7.55 & 6.51 & 20.90 \\
    & KAPING (Ours) & 33.39 & 11.36 & 33.08 & 20.92 & 87.44 & 17.19 & 20.2 & 45.77 \\
    \midrule
    \multirow{5}*{OPT (13B)} 
    & No Knowledge & 33.57 & 16.82 & 34.85 & 18.91 & 48.43 & 19.27 & 19.22 & 22.75 \\
    & Random Knowledge & 31.60 & 17.27 & 26.77 & 23.78 & 59.19 & 16.93 & 20.85 & 35.45 \\
    & Popular Knowledge & 22.98 & 13.64 & 24.49 & 18.34 & 59.64 & 11.72 & 12.05 & 30.69 \\
    & Generated Knowledge & 17.95 & 10.00 & 19.44 & 12.03 & 47.98 & 8.07 & 9.77 & 12.70 \\
    & KAPING (Ours) & 40.04 & 17.27 & 35.61 & 23.50 & 56.05 & 19.53 & 27.36 & 45.24 \\
    \midrule
    \multirow{5}*{T0 (3B)} 
    & No Knowledge & 13.82 & 10.00 & 14.39 & 10.89 & 49.33 & 14.06 & 8.79 & 7.94 \\
    & Random Knowledge & 19.57 & 9.09 & 15.66 & 12.32 & 58.30 & 8.59 & 9.77 & 6.88 \\
    & Popular Knowledge & 19.21 & 10.00 & 18.69 & 12.03 & 60.09 & 8.33 & 8.79 & 8.73 \\
    & Generated Knowledge & 13.11 & 11.36 & 12.63 & 12.61 & 54.71 & 12.50 & 10.10 & 3.70 \\
    & KAPING (Ours) & 29.98 & 10.45 & 26.01 & 12.32 & 55.16 & 12.24 & 11.40 & 10.85 \\
    \midrule
    \multirow{5}*{T0 (11B)} 
    & No Knowledge & 33.93 & 18.18 & 33.08 & 18.05 & 54.71 & 19.53 & 13.68 & 1.59 \\
    & Random Knowledge & 36.98 & 22.27 & 34.60 & 21.78 & 58.74 & 18.75 & 19.22 & 1.59 \\
    & Popular Knowledge & 38.42 & 24.09 & 38.64 & 24.36 & 58.74 & 17.45 & 18.57 & 1.06 \\
    & Generated Knowledge & 33.21 & 17.73 & 34.09 & 17.48 & 51.12 & 18.23 & 14.33 & 0.79 \\
    & KAPING (Ours) & 45.60 & 27.27 & 41.16 & 22.35 & 56.05 & 18.75 & 23.45 & 1.59 \\
    \midrule
    \multirow{5}*{GPT-3 (6.7B)} 
    & No Knowledge & 40.39 & 28.18 & 34.34 & 24.36 & 74.44 & 26.04 & 24.76 & 33.07 \\
    & Random Knowledge & 39.68 & 26.82 & 30.05 & 23.78 & 77.13 & 19.53 & 23.13 & 33.86 \\
    & Popular Knowledge & 40.57 & 25.00 & 32.83 & 22.64 & 70.85 & 21.35 & 21.17 & 31.48 \\
    & Generated Knowledge & 40.75 & 23.64 & 33.59 & 28.08 & 71.75 & 20.83 & 22.15 & 30.16 \\
    & KAPING (Ours) & 46.14 & 24.09 & 33.33 & 24.36 & 77.58 & 19.53 & 24.76 & 35.71 \\
    \midrule
    \multirow{5}*{GPT-3 (175B)} 
    & No Knowledge & 71.10 & 52.73 & 64.90 & 49.00 & 80.72 & 42.45 & 50.81 & 38.62 \\
    & Random Knowledge & 62.30 & 46.82 & 56.31 & 43.55 & 86.10 & 38.54 & 48.21 & 36.51 \\
    & Popular Knowledge & 68.40 & 54.09 & 58.84 & 46.42 & 81.61 & 37.76 & 47.88 & 33.60 \\
    & Generated Knowledge & 70.56 & 56.82 & 64.14 & 48.14 & 85.65 & 44.79 & 49.19 & 29.63 \\
    & KAPING (Ours) & 69.48 & 56.36 & 63.89 & 48.14 & 82.96 & 45.57 & 49.84 & 41.01 \\
    \midrule
    \multirow{5}*{\textbf{Average}} 
    & No Knowledge & 27.41 & 16.14 & 26.36 & 17.74 & 65.79 & 17.81 & 16.58 & 18.81 \\
    & Random Knowledge & 29.23 & 16.91 & 24.04 & 20.37 & \textbf{72.47} & 15.70 & 17.56 & 27.01 \\
    & Popular Knowledge & 28.85 & 17.41 & 25.76 & 20.09 & 72.20 & 15.05 & 16.61 & 24.79 \\
    & Generated Knowledge & 21.85 & 14.41 & 21.06 & 16.28 & 65.38 & 12.82 & 12.48 & 18.39 \\
    & KAPING (Ours) & \textbf{36.70} & \textbf{19.00} & \textbf{32.30} & \textbf{21.44} & 70.58 & \textbf{18.10} & \textbf{20.49} & \textbf{31.88} \\
    \bottomrule
    \end{tabular}
    }
    \end{center}
    \vspace{-0.125in}
    \caption{\small \textbf{Results across question types} -- Generic, Multihop, Intersection, Difference, Comparative, Superlative, Ordinal, and Count -- in the Mintaka dataset with various LLMs, where we report the generation accuracy. The number inside the parentheses in the first row denotes the number of data instances per category. We provide averaged results in the last row, namely \textbf{Average}.}
    \label{tab:appendix:complexity_type}
\end{table*}

\onecolumn

\begin{center}

\begingroup

\fontsize{7pt}{11pt}\selectfont

\begin{longtable}{lp{1.0in}p{1.2in}p{1.5in}p{1.5in}}
\caption{\textbf{Generation examples of GPT-3 (6.7B)} on the WebQSP w/ Freebase dataset, where we randomly sample 30 examples.} 
\vspace{-0.1in}
\label{tab:examples_gpt3} \\

\hline 
\textbf{Index} & \textbf{Question} & \textbf{Answer Entities} & \textbf{No Knowledge} & \textbf{KAPING} 
\\ \hline 
\endfirsthead

\multicolumn{5}{c}%
{{\bfseries \tablename\ \thetable{} -- Continued from the previous page}} \\
\hline 
\textbf{Index} & \textbf{Question} & \textbf{Answer Entities} & \textbf{No Knowledge} & \textbf{KAPING} 
\\ \hline 
\endhead 

\multicolumn{5}{r}{\textit{\textbf{Continued on the next page}}} \\ \hline
\endfoot

\hline \hline
\endlastfoot

1&
who is tatyana ali parents?&
[Sheriff Ali, Sonia Ali]&
Tatyana Ali's parents are unknown.&
Tatyana Ali's parents are Sheriff Ali and Sonia Ali.\\

\noalign{\vskip 0.25ex}\cdashline{1-5}\noalign{\vskip 0.75ex}

2&
what language do most australians speak?&
[English Language, Esperanto Language, Lojban]&
Most Australians speak English.&
Most Australians speak English.\\

\noalign{\vskip 0.25ex}\cdashline{1-5}\noalign{\vskip 0.75ex}

3&
who played darth vader in the first movie?&
[David Prowse, James Earl Jones]&
James Earl Jones&
Mark Hamill\\

\noalign{\vskip 0.25ex}\cdashline{1-5}\noalign{\vskip 0.75ex}

4&
who does michael keaton play in cars?&
[Chick Hicks, Chick Hicks]&
Michael Keaton plays the role of Larry "Lester" Fenton in Cars.&
The role of Michael Keaton in Cars is that of a voice actor.\\

\noalign{\vskip 0.25ex}\cdashline{1-5}\noalign{\vskip 0.75ex}

5&
what was pink floyd's first album?&
[Psychedelic Session-compilation album]&
The first album released by Pink Floyd was the self-titled "Pink Floyd" in 1965.&
Pink Floyd's first album was called "Pink Floyd".\\

\noalign{\vskip 0.25ex}\cdashline{1-5}\noalign{\vskip 0.75ex}

6&
where is walton county ga?&
[Atlanta-Sandy Springs-Marietta, GA Metropolitan Statistical Area, Georgia]&
Walton County is located in Georgia.&
According to the US Census 2000, Walton County is located in between 30641 and 30025, in Gratis, Georgia.\\

\noalign{\vskip 0.25ex}\cdashline{1-5}\noalign{\vskip 0.75ex}

7&
what kind government does egypt have?&
[Semi-presidential system, Provisional government]&
Egypt has a presidential system.&
Egpyt has a semi-presidential system of government.\\

\noalign{\vskip 0.25ex}\cdashline{1-5}\noalign{\vskip 0.75ex}

8&
where is english spoken around the world?&
[Bahamas, Bangladesh, ..., United Kingdom, United States of America, Australia, New Zealand, Canada]&
English is spoken around the world.&
English is spoken in different countries around the world. England, United States of America, United Kingdom\\

\noalign{\vskip 0.25ex}\cdashline{1-5}\noalign{\vskip 0.75ex}

9&
who did steve spurrier play pro football for?&
[San Francisco 49ers, Tampa Bay Buccaneers]&
Steve Spurrier played professional football for the Tampa Bay Buccaneers from 1976 to 1978.&
St. Louis Rams\\

\noalign{\vskip 0.25ex}\cdashline{1-5}\noalign{\vskip 0.75ex}

10&
what kind of music did franz liszt compose?&
[Opera, Romantic music, Art song, Classical music, Chamber music]&
Franz Liszt composed a great deal of classical music, including symphonies, operas, and piano sonatas.&
Franz Liszt composed both classical and popular music. Some of his most famous works include the opera Don Giovanni, the symphonies Annabella and Egmont, and the song Totentanz.\\

\noalign{\vskip 0.25ex}\cdashline{1-5}\noalign{\vskip 0.75ex}

11&
what was the currency used in france before the euro?&
[Assignat, French franc]&
Until the late 1990s, France used the franc as its official currency.&
the currency used in France before the euro was the French franc.\\

\noalign{\vskip 0.25ex}\cdashline{1-5}\noalign{\vskip 0.75ex}

12&
where is the riviera casino?&
[Las Vegas]&
The Riviera Casino is located in Las Vegas, NV.&
Riviera casino is located on the Las Vegas strip in Nevada.\\

\noalign{\vskip 0.25ex}\cdashline{1-5}\noalign{\vskip 0.75ex}

13&
what does wh smith stand for?&
[WHSmith]&
Whit Smith is an American printing and publishing company. It exclusively publishes books for children and young adults.&
William Henry Smith \\

\noalign{\vskip 0.25ex}\cdashline{1-5}\noalign{\vskip 0.75ex}

14&
who will play mr gray in the film?&
[Jamie Dornan]&
Mr. Gray has not been announced, but it is likely that the role will be played by a younger actor.&
The film will be based on the book "Fifty Shades of Grey" by E.L. James, and the character Mr. Grey will be played by Jamie Dornan.\\

\noalign{\vskip 0.25ex}\cdashline{1-5}\noalign{\vskip 0.75ex}

15&
what religions are popular in france?&
[Catholicism]&
There are a number of religions popular in France, some of which are Christianity, Islam, Hinduism, and Judaism.&
According to the 2016 census, 51.6\% of the population of France identify as Christians, including 35.9\% who identify as Catholic and 14.4\% as Protestant. The remaining 5.1\% of the population identify with other religions. Furthermore, according to the 2012 census, French is the language most commonly spoken at home by the population of France. It is also the language most commonly spoken in the administrative regions of metropolitan France, with Italian second and French Creole third.\\

\noalign{\vskip 0.25ex}\cdashline{1-5}\noalign{\vskip 0.75ex}

16&
who did nasri play for before arsenal?&
[Olympique de Marseille]&
he played for Universidad de Chile and Monaco before joining Arsenal in 2011.&
Samir Nasri played for different teams before Arsenal, including Manchester City and Marseille.\\

\noalign{\vskip 0.25ex}\cdashline{1-5}\noalign{\vskip 0.75ex}

17&
who plays ziva david on ncis?&
[Cote de Pablo]&
Ziva David is played by Sandra Oh.&
Ziva David is played by actress Debra Messing in the NBC television series NCIS.\\

\noalign{\vskip 0.25ex}\cdashline{1-5}\noalign{\vskip 0.75ex}

18&
where did adolf hitler die?&
[Berlin]&
Adolf Hitler died in his home town of Braunau am Inn, Austria on April 30, 1945, days before German forces unconditionally surrendered in World War II.&
Adolf Hitler committed suicide in Berlin, Germany.\\

\noalign{\vskip 0.25ex}\cdashline{1-5}\noalign{\vskip 0.75ex}

19&
what was hitler the leader of?&
[Hitler Youth, Gestapo, 1st SS Panzer Division Leibstandarte SS Adolf Hitler, Nazi Party, Schutzstaffel, Waffen-SS, Wehrmacht, Sturmabteilung]&
Hitler was the leader of Nazi Germany, which was a fascist dictatorship in Europe during the 1930s and 1940s.&
The leader of Nazi Germany was Adolf Hitler.\\

\noalign{\vskip 0.25ex}\cdashline{1-5}\noalign{\vskip 0.75ex}

20&
what types of art did andy warhol do?&
[Modern art, Pop art]&
Andy Warhol did a variety of art, including paintings, sculptures, and prints.&	visual\_art.visual\_artist.artworks, men in her life\\

\noalign{\vskip 0.25ex}\cdashline{1-5}\noalign{\vskip 0.75ex}

21&
who owns the portland press herald?&
[Blethen Maine Newspapers, Inc.]&
The Portland Press Herald is not owned by any specific individual or entity. It is an independent, locally owned newspaper.&
The Portland Press Herald is owned by the Blethen Maine Newspapers, Inc.\\

\noalign{\vskip 0.25ex}\cdashline{1-5}\noalign{\vskip 0.75ex}

22&
where was the vietnam war location?&
[Vietnam, Cambodia, Laos, Southeast Asia, South Vietnam, North Vietnam]&
The Vietnam War was fought in Vietnam, Cambodia, and Laos
&The Vietnam War was fought in Cambodia, Laos, North Vietnam, and South Vietnam.\\

\noalign{\vskip 0.25ex}\cdashline{1-5}\noalign{\vskip 0.75ex}

23&
who was isaac newton and what did he do?&
[Chemist, Philosopher, Mathematician, Physicist, Scientist, Astronomer]&
Isaac Newton was a physicist, mathematician, astronomer, and author of the Newtonian Theory of Motion. He is most famous for his three laws of motion and his theory of universal gravitation.&
Isaac Newton was a scientist, mathematician, and philosopher. \\

\noalign{\vskip 0.25ex}\cdashline{1-5}\noalign{\vskip 0.75ex}

24&
what standard time is michigan?&
[UTC-05:00, Central Time Zone, Eastern Time Zone]&
Daylight Saving Time is in effect in Michigan from the first Sunday in April to the last Sunday in October.&
Michigan is in the Eastern Time Zone.\\

\noalign{\vskip 0.25ex}\cdashline{1-5}\noalign{\vskip 0.75ex}

25&
what is the kentucky state bird?&
[Cardinal]&
The Kentucky state bird is the cardinal.&
The kentucky state bird is the pigeon.\\

\noalign{\vskip 0.25ex}\cdashline{1-5}\noalign{\vskip 0.75ex}

26&
what do people in jamaica speak?&
[Jamaican English, Jamaican Creole English Language, Jamaican English]&
people in Jamaica speak English.&
Jamaican Creole English Language\\

\noalign{\vskip 0.25ex}\cdashline{1-5}\noalign{\vskip 0.75ex}

27&
who wrote the jana gana mana?&
[Ram Singh Thakur, Rabindranath Tagore]&
The author of the Jana Gana Mana is unknown.&
Jana Gana Mana was written by the composer Ram Singh Thakur.\\

\noalign{\vskip 0.25ex}\cdashline{1-5}\noalign{\vskip 0.75ex}

28&
what is the currency of the czech republic?&
[Czech koruna]&
The Czech Republic currency is the Czech koruna.&
The currency of the Czech Republic is the Czech koruna.\\

\noalign{\vskip 0.25ex}\cdashline{1-5}\noalign{\vskip 0.75ex}

29&
what highschool did harper lee go to?&
[Monroe County High School]&
Harper Lee attended the Monroeville, Alabama high school.&
Harper Lee attended Monroeville High School.\\

\noalign{\vskip 0.25ex}\cdashline{1-5}\noalign{\vskip 0.75ex}

30&
what did michael crabtree do?&
[Wide receiver, Athlete, American football player, Wide receiver]&
Michael Crabtree is a former football player who played for the Baltimore Ravens. Crabtree caught a pass in the end zone to tie the game against the San Francisco 49ers in the final seconds of the game in 2013.&
He is an American football player.

\end{longtable}

\endgroup

\end{center}

\twocolumn

\onecolumn

\begin{center}

\begingroup

\fontsize{7pt}{11pt}\selectfont

\begin{longtable}{lp{1.0in}p{1.2in}p{1.5in}p{1.5in}}
\caption{\textbf{Generation examples of T0 (3B)} on the WebQSP w/ Freebase dataset, where we randomly sample 30 examples.} 
\vspace{-0.1in}
\label{tab:examples_t0} \\

\hline 
\textbf{Index} & \textbf{Question} & \textbf{Answer Entities} & \textbf{No Knowledge} & \textbf{KAPING} 
\\ \hline 
\endfirsthead

\multicolumn{5}{c}%
{{\bfseries \tablename\ \thetable{} -- Continued from the previous page}} \\
\hline 
\textbf{Index} & \textbf{Question} & \textbf{Answer Entities} & \textbf{No Knowledge} & \textbf{KAPING} 
\\ \hline 
\endhead 

\multicolumn{5}{r}{\textit{\textbf{Continued on the next page}}} \\ \hline
\endfoot

\hline \hline
\endlastfoot

1&
who is tatyana ali parents?&
[Sheriff Ali, Sonia Ali]&
ali is the daughter of ali khan and tatyana ali&
Sheriff Ali\\

\noalign{\vskip 0.25ex}\cdashline{1-5}\noalign{\vskip 0.75ex}

2&
what language do most australians speak?&
[English Language, Esperanto Language, Lojban]&
Australian English&
English Language\\

\noalign{\vskip 0.25ex}\cdashline{1-5}\noalign{\vskip 0.75ex}

3&
who played darth vader in the first movie?&
[David Prowse, James Earl Jones]&
Darth Vader&
George Lucas\\

\noalign{\vskip 0.25ex}\cdashline{1-5}\noalign{\vskip 0.75ex}

4&
who does michael keaton play in cars?&
[Chick Hicks, Chick Hicks]&
a narrator&
George A. Douglas\\

\noalign{\vskip 0.25ex}\cdashline{1-5}\noalign{\vskip 0.75ex}

5&
what was pink floyd's first album?&
[Psychedelic Session-compilation album]&
The Pink Floyd album&
Pink Floyd\\

\noalign{\vskip 0.25ex}\cdashline{1-5}\noalign{\vskip 0.75ex}

6&
where is walton county ga?&
[Atlanta-Sandy Springs-Marietta, GA Metropolitan Statistical Area, Georgia]&
Walton County is a county in the U.S. state of Ohio.&
Walton County, Georgia\\

\noalign{\vskip 0.25ex}\cdashline{1-5}\noalign{\vskip 0.75ex}

7&
what kind government does egypt have?&
[Semi-presidential system, Provisional government]&
government&
Semi-presidential system\\

\noalign{\vskip 0.25ex}\cdashline{1-5}\noalign{\vskip 0.75ex}

8&
where is english spoken around the world?&
[Bahamas, Bangladesh, ..., United Kingdom, United States of America, Australia, New Zealand, Canada]&
The English language is spoken by approximately 6.1 million people in the United Kingdom, with the remainder in the United States, Canada, Australia, New Zealand, Canada, and the United States Virgin Islands.&
England\\

\noalign{\vskip 0.25ex}\cdashline{1-5}\noalign{\vskip 0.75ex}

9&
who did steve spurrier play pro football for?&
[San Francisco 49ers, Tampa Bay Buccaneers]&
the New York Giants&
the New York Giants\\

\noalign{\vskip 0.25ex}\cdashline{1-5}\noalign{\vskip 0.75ex}

10&
what kind of music did franz liszt compose?&
[Opera, Romantic music, Art song, Classical music, Chamber music]&
classical&
Art song\\

\noalign{\vskip 0.25ex}\cdashline{1-5}\noalign{\vskip 0.75ex}

11&
what was the currency used in france before the euro?&
[Assignat, French franc]&
The pound was used in France before the euro.&
French franc\\

\noalign{\vskip 0.25ex}\cdashline{1-5}\noalign{\vskip 0.75ex}

12&
where is the riviera casino?&
[Las Vegas]&
Riviera Maya, Mexico&
Las Vegas\\

\noalign{\vskip 0.25ex}\cdashline{1-5}\noalign{\vskip 0.75ex}

13&
what does wh smith stand for?&
[WHSmith]&
smith is a scottish slang word for a smith&
William Henry Smith\\

\noalign{\vskip 0.25ex}\cdashline{1-5}\noalign{\vskip 0.75ex}

14&
who will play mr gray in the film?&
[Jamie Dornan]&
Corey Haim&
Christian Grey\\

\noalign{\vskip 0.25ex}\cdashline{1-5}\noalign{\vskip 0.75ex}

15&
what religions are popular in france?&
[Catholicism]&
Roman Catholicism&
Protestantism\\

\noalign{\vskip 0.25ex}\cdashline{1-5}\noalign{\vskip 0.75ex}

16&
who did nasri play for before arsenal?&
[Olympique de Marseille]&
Arsenal&
Arsenal\\

\noalign{\vskip 0.25ex}\cdashline{1-5}\noalign{\vskip 0.75ex}

17&
who plays ziva david on ncis?&
[Cote de Pablo]&
sarah savage&
Donald P. Bellisario\\

\noalign{\vskip 0.25ex}\cdashline{1-5}\noalign{\vskip 0.75ex}

18&
where did adolf hitler die?&
[Berlin]&
in a concentration camp&
Berlin\\

\noalign{\vskip 0.25ex}\cdashline{1-5}\noalign{\vskip 0.75ex}

19&
what was hitler the leader of?&
[Hitler Youth, Gestapo, 1st SS Panzer Division Leibstandarte SS Adolf Hitler, Nazi Party, Schutzstaffel, Waffen-SS, Wehrmacht, Sturmabteilung]&
the Nazi Party&
World War I\\

\noalign{\vskip 0.25ex}\cdashline{1-5}\noalign{\vskip 0.75ex}

20&
what types of art did andy warhol do?&
[Modern art, Pop art]&
art&
visual art.visual\_artist.artworks, Men in Her Life\\

\noalign{\vskip 0.25ex}\cdashline{1-5}\noalign{\vskip 0.75ex}

21&
who owns the portland press herald?&
[Blethen Maine Newspapers, Inc.]&
The Portland Press Herald is a local, independent newspaper based in Portland, Oregon, United States.&
Blethen Maine Newspapers\\

\noalign{\vskip 0.25ex}\cdashline{1-5}\noalign{\vskip 0.75ex}

22&
where was the vietnam war location?&
[Vietnam, Cambodia, Laos, Southeast Asia, South Vietnam, North Vietnam]&
Vietnam War&
Vietnam\\

\noalign{\vskip 0.25ex}\cdashline{1-5}\noalign{\vskip 0.75ex}

23&
who was isaac newton and what did he do?&
[Chemist, Philosopher, Mathematician, Physicist, Scientist, Astronomer]&
was a british physicist&
physicist\\

\noalign{\vskip 0.25ex}\cdashline{1-5}\noalign{\vskip 0.75ex}

24&
what standard time is michigan?&
[UTC-05:00, Central Time Zone, Eastern Time Zone]&
Standard time is local time in the United States.&
Midwestern United States\\

\noalign{\vskip 0.25ex}\cdashline{1-5}\noalign{\vskip 0.75ex}

25&
what is the kentucky state bird?&
[Cardinal]&
a red-winged blackbird&
Pigeon\\

\noalign{\vskip 0.25ex}\cdashline{1-5}\noalign{\vskip 0.75ex}

26&
what do people in jamaica speak?&
[Jamaican English, Jamaican Creole English Language, Jamaican English]&
People in Jamaica speak English&
Jamaican Creole English Language\\

\noalign{\vskip 0.25ex}\cdashline{1-5}\noalign{\vskip 0.75ex}

27&
who wrote the jana gana mana?&
[Ram Singh Thakur, Rabindranath Tagore]&
the author of the jana gana mana is a poet&
Rabindranath Tagore\\

\noalign{\vskip 0.25ex}\cdashline{1-5}\noalign{\vskip 0.75ex}

28&
what is the currency of the czech republic?&
[Czech koruna]&
the czech koruna&
Czech koruna\\

\noalign{\vskip 0.25ex}\cdashline{1-5}\noalign{\vskip 0.75ex}

29&
what highschool did harper lee go to?&
[Monroe County High School]&
Harper Lee High School&
Monroeville High School\\

\noalign{\vskip 0.25ex}\cdashline{1-5}\noalign{\vskip 0.75ex}

30&
what did michael crabtree do?&
[Wide receiver, Athlete, American football player, Wide receiver]&
did not have a job&
American football player\\

\end{longtable}

\endgroup

\end{center}

\twocolumn

\onecolumn

\begin{center}

\begingroup

\fontsize{7pt}{11pt}\selectfont

\begin{longtable}{lp{0.8in}p{2.1in}p{0.7in}p{1.6in}}
\caption{\textbf{Generation examples of our KAPING framework} with four different models: T5 (11B), OPT (13B), T0 (11B), and GPT-3 (175B) on the WebQSP w/ Wikidata dataset. Note that the answer entities are labeled entities in the test dataset.} 
\vspace{-0.1in}
\label{tab:examples_ours} \\

\hline 
\textbf{Index} & \textbf{Question} & \textbf{Retrieved Triples} & \textbf{Answer Entities} & \textbf{Generated Answers} 
\\ \hline 
\endfirsthead

\multicolumn{5}{c}%
{{\bfseries \tablename\ \thetable{} -- Continued from the previous page}} \\
\hline 
\textbf{Index} & \textbf{Question} & \textbf{Retrieved Triples} & \textbf{Answer Entities} & \textbf{Generated Answers} 
\\ \hline 
\endhead 

\multicolumn{5}{r}{\textit{\textbf{Continued on the next page}}} \\ \hline
\endfoot

\hline \hline
\endlastfoot

\multirow{4}{*}{1}&
\multirow{4}{0.8in}{
what is the name of the currency used in china?
}&
\multirow{4}{2.1in}{
(People's Republic of China, currency, renminbi),
(People's Republic of China, short name, text: Chine),
(People's Republic of China, short name, text: Chiny),
(People's Republic of China, language used, Chinese),
(People's Republic of China, central bank, People's Bank of China),
(People's Republic of China, language used, China Buriat),
(People's Republic of China, demonym, text: Chinesin),
(People's Republic of China, language used, Jingpho),
(People's Republic of China, language used, Langnian Buyang),
(People's Republic of China, language used, Then)
}&
\multirow{4}{0.7in}{
[Renminbi]
}&
\textbf{T5}: Renminbi \\ \\
& & & & 
\textbf{OPT}: Renminbi \\ \\
& & & & 
\textbf{T0}: renminbi \\ \\
& & & & 
\textbf{GPT-3}: The name of the currency used in china is renminbi.
\\ \\ \\ \\ 

\noalign{\vskip 0.25ex}\cdashline{1-5}\noalign{\vskip 0.75ex}

\multirow{4}{*}{2}&
\multirow{4}{0.8in}{
what does george wilson do for a living?
}&
\multirow{4}{2.1in}{
(George Wilson, country of citizenship, United States of America),
(George Wilson, occupation, American football player),
(George Wilson, sex or gender, male),
(George Wilson, given name, George),
(George Wilson, instance of, human),
(George Wilson, place of birth, Paducah),
(George Wilson, member of sports team, Arkansas Razorbacks football),
(George Wilson, member of sports team, Detroit Lions),
(George Wilson, member of sports team, Buffalo Bills),
(George Wilson, educated at, Paducah Tilghman High School)
}&
\multirow{4}{0.7in}{
[American football player]
}&
\textbf{T5}: Wilson, George was born on August 10, 1972 in Paducah, Kentucky. He is an American football player. \\ \\
& & & & 
\textbf{OPT}: George Wilson is an American football player. \\ \\
& & & & 
\textbf{T0}: American football player \\ \\
& & & & 
\textbf{GPT-3}: George Wilson is an American football player.
\\ 

\noalign{\vskip 0.25ex}\cdashline{1-5}\noalign{\vskip 0.75ex}

\multirow{4}{*}{3}&
\multirow{4}{0.8in}{
what was augustus caesar famous for?
}&
\multirow{4}{2.1in}{
(Augustus, Roman cognomen, Caesar),
(Augustus, father, Julius Caesar),
(Augustus, occupation, Ancient Roman politician),
(Augustus, child, Lucius Caesar),
(Augustus, child, Gaius Caesar),
(Augustus, Roman cognomen, Caepias),
(Augustus, relative, Julius Caesar),
(Augustus, Roman praenomen, Gaius),
(Augustus, father, Gaius Octavius),
(Augustus, family, Julii Caesares)
}&
\multirow{4}{0.7in}{
[Roman emperor, Politician]
}&
\textbf{T5}: Augustus was a Roman politician who ruled from 14 BC to 14 AD. He was the son of Gaius Octavius and Octavia. He was the first emperor of the Roman Empire. \\ \\
& & & & 
\textbf{OPT}: Augustus Caesar was famous for being the first Roman emperor. \\ \\
& & & & 
\textbf{T0}: He was the first Roman Emperor \\ \\
& & & & 
\textbf{GPT-3}: Augustus Caesar was famous for being the first Roman Emperor.
\\ 

\noalign{\vskip 0.25ex}\cdashline{1-5}\noalign{\vskip 0.75ex}

\multirow{4}{*}{4}&
\multirow{4}{0.8in}{
what document did james madison write?
}&
\multirow{4}{2.1in}{
(James Madison, described by source, Obalky knih.cz),
(James Madison, described by source, The Nuttall Encyclopædia),
(James Madison, Commons Creator page, text: James Madison),
(James Madison, writing language, English),
(James Madison, described by source, Brockhaus and Efron Encyclopedic Dictionary),
(James Madison, notable work, Federalist Papers),
(James Madison, occupation, statesperson),
(James Madison, occupation, writer),
(James Madison, described by source, Library of the World's Best Literature),
(James Madison, position held, President of the United States)
}&
\multirow{4}{0.7in}{
[Federalist Papers]
}&
\textbf{T5}: federalist papers \\ \\
& & & & 
\textbf{OPT}: I think the answer is "The Federalist Papers" \\ \\
& & & & 
\textbf{T0}: Federalist Papers \\ \\
& & & & 
\textbf{GPT-3}: the Federalist Papers
\\ \\ \\ \\ \\

\noalign{\vskip 0.25ex}\cdashline{1-5}\noalign{\vskip 0.75ex}

\multirow{4}{*}{5}&
\multirow{4}{0.8in}{
where did george michael go to school?
}&
\multirow{4}{2.1in}{
(George Michael, educated at, Bushey Meads School),
(George Michael, occupation, singer),
(George Michael, occupation, singer-songwriter),
(George Michael, place of birth, East Finchley),
(George Michael, occupation, record producer),
(George Michael, occupation, multi-instrumentalist),
(George Michael, occupation, recording artist),
(George Michael, occupation, composer),
(George Michael, occupation, songwriter),
(George Michael, country of citizenship, United Kingdom)
}&
\multirow{4}{0.7in}{
[Bushey Meads School]
}&
\textbf{T5}: . Question: where did george michael go to school? Answer: bushey meads school \\ \\
& & & & 
\textbf{OPT}: Bushey Meads School \\ \\
& & & & 
\textbf{T0}: Bushey Meads School \\ \\
& & & & 
\textbf{GPT-3}: Bushey Meads School
\\ \\ 

\noalign{\vskip 0.25ex}\cdashline{1-5}\noalign{\vskip 0.75ex}

\multirow{4}{*}{6}&
\multirow{4}{0.8in}{
what college did joakim noah?
}&
\multirow{4}{2.1in}{
(Joakim Noah, educated at, University of Florida),
(Joakim Noah, educated at, Lawrenceville School),
(Joakim Noah, educated at, United Nations International School),
(Joakim Noah, educated at, Poly Prep Country Day School),
(Joakim Noah, sport, basketball),
(Joakim Noah, league, National Basketball Association),
(Joakim Noah, league, NCAA Division I men's basketball),
(Joakim Noah, member of sports team, Florida Gators men's basketball),
(Joakim Noah, member of sports team, Florida Gators men's basketball),
(Joakim Noah, league, NBA G League)
}&
\multirow{4}{0.7in}{
[University of Florida]
}&
\textbf{T5}: Joakim Noah, college, University of Florida \\ \\
& & & & 
\textbf{OPT}: University of Florida Question: what sport did joakim noah? Answer: basketball Question: what league did joakim noah? Answer: NBA \\ \\
& & & & 
\textbf{T0}: University of Florida \\ \\
& & & & 
\textbf{GPT-3}: Joakim Noah attended the University of Florida.
\\

\noalign{\vskip 0.25ex}\cdashline{1-5}\noalign{\vskip 0.75ex}

\multirow{4}{*}{7}&
\multirow{4}{0.8in}{
where does tim cook work?
}&
\multirow{4}{2.1in}{
(Tim Cook, occupation, businessperson),
(Tim Cook, occupation, engineer),
(Tim Cook, occupation, computer scientist),
(Tim Cook, place of birth, Mobile),
(Tim Cook, occupation, chief executive officer),
(Tim Cook, described by source, Lentapedia),
(Tim Cook, country of citizenship, United States of America),
(Tim Cook, employer, Apple Inc.),
(Tim Cook, residence, Palo Alto),
(Tim Cook, educated at, Fuqua School of Business)
}&
\multirow{4}{0.7in}{
[Apple Inc.]
}&
\textbf{T5}: Tim Cook, educated at, Fuqua School of Business Question: where does tim cook work? Answer: Tim Cook, residence, Palo Alto Question: where does tim cook work? Answer: Tim Cook, employer, Apple Inc. \\ \\
& & & & 
\textbf{OPT}: I think the answer is "Apple Inc." \\ \\
& & & & 
\textbf{T0}: Apple Inc. \\ \\
& & & & 
\textbf{GPT-3}: Tim Cook works at Apple Inc.
\\

\noalign{\vskip 0.25ex}\cdashline{1-5}\noalign{\vskip 0.75ex}

\multirow{4}{*}{8}&
\multirow{4}{0.8in}{
what is the political system of south africa?
}&
\multirow{4}{2.1in}{
(South Africa, basic form of government, representative democracy),
(South Africa, basic form of government, parliamentary republic),
(South Africa, Democracy Index, quantity: +7.05 - unit: 1),
(South Africa, Democracy Index, quantity: +7.24 - unit: 1),
(South Africa, Democracy Index, quantity: +7.24 - unit: 1),
(South Africa, Democracy Index, quantity: +7.24 - unit: 1),
(South Africa, legislative body, Parliament of South Africa),
(South Africa, executive body, Government of South Africa),
(South Africa, office held by head of government, State President of South Africa),
(South Africa, member of, Southern African Development Community)
}&
\multirow{4}{0.7in}{
[Constitutional republic, Parliamentary republic]
}&
\textbf{T5}: Question: what is the political system of south africa? Answer: Answer: parliamentary republic \\ \\
& & & & 
\textbf{OPT}: South Africa is a parliamentary republic. Question: what is the capital of south africa? Answer: Pretoria is the capital of South Africa. Question: what is the currency of south africa? Answer: South African Rand. Question: what is the population of south africa? Answer: South Africa has a population of about 50 million. Question: what is the official language of south africa? Answer: Afrikaans is the official language of South Africa. \\ \\
& & & & 
\textbf{T0}: parliamentary republic \\ \\
& & & & 
\textbf{GPT-3}: The political system of South Africa is a parliamentary republic.
\\

\noalign{\vskip 0.25ex}\cdashline{1-5}\noalign{\vskip 0.75ex}

\multirow{4}{*}{9}&
\multirow{4}{0.8in}{
what record label is kelly clarkson signed with?
}&
\multirow{4}{2.1in}{
(Kelly Clarkson, record label, RCA Records),
(Kelly Clarkson, record label, Atlantic Records),
(Kelly Clarkson, record label, Syco Music),
(Kelly Clarkson, record label, 19 Recordings),
(Kelly Clarkson, occupation, record producer),
(Kelly Clarkson, occupation, recording artist),
(Kelly Clarkson, occupation, musician),
(Kelly Clarkson, occupation, singer),
(Kelly Clarkson, genre, pop music),
(Kelly Clarkson, given name, Kelly)
}&
\multirow{4}{0.7in}{
[RCA Records, Sony BMG, Sony Music Entertainment, 19 Recordings]
}&
\textbf{T5}: Kelly Clarkson is a singer, songwriter, and actress. She is signed to RCA Records, a division of RCA Records. \\ \\
& & & & 
\textbf{OPT}: 19 Recordings Question: what is kelly clarkson's first name? Answer: kelly Question: what is kelly clarkson's last name? Answer: clarkson \\ \\
& & & & 
\textbf{T0}: RCA Records \\ \\
& & & & 
\textbf{GPT-3}: RCA Records
\\

\noalign{\vskip 0.25ex}\cdashline{1-5}\noalign{\vskip 0.75ex}

\multirow{4}{*}{10}&
\multirow{4}{0.8in}{
what kind of products does nike sell?
}&
\multirow{4}{2.1in}{
(Nike, product or material produced, sports equipment),
(Nike, product or material produced, clothing),
(Nike, product or material produced, footwear),
(Nike, industry, retail),
(Nike, industry, clothing industry),
(Nike, instance of, brand),
(Nike, subsidiary, Air Jordan product line),
(Nike, industry, industry),
(Nike, Commons category, text: Nike, Inc.),
(Nike, subsidiary, Converse)
}&
\multirow{4}{0.7in}{
[Athletic Shoes \& Apparel, Footwear Manufacturing, Fashion accessory, Sports equipment, Clothing, Footwear]
}&
\textbf{T5}: Nike, Inc. is a multinational corporation headquartered in Beaverton, Oregon, United States. It is the world's largest manufacturer of athletic shoes and apparel, and the second-largest manufacturer of sports equipment. It is the world's largest sportswear manufacturer, and the world's largest manufacturer of footwear for men, women and children. \\ \\
& & & & 
\textbf{OPT}: shoes \\ \\
& & & & 
\textbf{T0}: Footwear \\ \\
& & & & 
\textbf{GPT-3}: Nike sells products in the categories of footwear, clothing, and sports equipment.
\\

\end{longtable}

\endgroup

\end{center}

\twocolumn

\end{document}